# RoadAnnoter: An AI-Assisted Desktop Tool for Semantic Segmentation Annotation of Road-Scene Images

**Nazmul Haque, PhD***
Assistant Professor
Accident Research Institute (ARI)
Bangladesh University of Engineering and Technology (BUET)
Email: nhaque@ari.buet.ac.bd
ORCID: *https://orcid.org/0009-0006-4698-3366*

**Arnob Protim Roy**
Graduate Research Assistant
Department of Civil Engineering (CE)
Bangladesh University of Engineering and Technology (BUET)
Email: pr.arnob@gmail.com

***Corresponding Author**

# RoadAnnoter: An AI-Assisted Desktop Tool for Semantic Segmentation Annotation of Road-Scene Images

**Abstract**

Accurate semantic segmentation of road-scene images is essential for data-driven road safety research; however, polygon-level annotation remains labor-intensive and limits the scalability of labelled dataset development. This study presents RoadAnnoter v1.0, an open-source desktop application integrating six complementary human-in-the-loop annotation modes within a unified Python-based environment: manual polygon delineation, SAM Prompt, Smart Select, Edge Fill, pretrained semantic segmentation using Cityscapes-trained SegFormer models, and task-specific segmentation using TinySegNet, a lightweight MobileNetV3-Small-based network trainable with accumulated annotator-generated labels. SAM-based modes employ promptable segmentation using SAM and, while Edge Fill combines Canny edge detection with colour-similarity-based flood filling in the CIELAB colour space. The pretrained models generate semantic suggestions for common road-scene categories, whereas TinySegNet supports adaptation to project-specific features through iterative training. All six modes employ a shared mask-to-polygon conversion pipeline incorporating morphological processing, Gaussian smoothing, contour extraction, and Douglas–Peucker simplification, with annotations exported in the YOLO segmentation format. The application was deployed to annotate Google Street View images collected at recorded road accident locations in Bangladesh for a vision-language model-based investigation of free-flow speed estimation. Qualitative assessment across representative road-scene scenarios demonstrated the complementary applicability of the annotation modes to diverse feature geometries. Visual comparison with Roboflow illustrated similar qualitative boundary characteristics for principal road-scene regions, although no quantitative accuracy assessment was conducted. RoadAnnoter provides an integrated, locally operated annotation framework for developing road-scene segmentation datasets in transportation engineering and road safety research.

# 1 Introduction

Road traffic crashes result in approximately 1.19 million fatalities annually, representing the eighth leading cause of death globally and the leading cause of mortality among individuals aged 5–29 years, with more than 90% of these fatalities occurring in low- and middle-income countries (WHO, 2023). In addition to their immediate human consequences, road traffic crashes impose substantial macroeconomic costs, estimated at 3–5% of gross domestic product (GDP), through medical expenditures, infrastructure damage, and productivity losses (WHO, 2023). Addressing this global road safety challenge requires systematic and scalable approaches to characterizing the road environment, including roadway geometry, pavement conditions, roadside hazards, and traffic control devices, using street-view imagery (Naila et al., 2026). Such characterization provides essential spatial information for investigating the relationship between road infrastructure characteristics and traffic safety outcomes.

Recent advances in deep learning have substantially enhanced the automated interpretation of road-scene imagery. Semantic segmentation architectures, evolving from fully convolutional networks (Long et al., 2015) to atrous-convolution-based architectures such as DeepLabV3+ (Chen et al., 2018) and hierarchical vision transformer architectures such as SegFormer (Xie et al., 2021), have demonstrated strong capabilities in delineating road surfaces, vehicles, infrastructure, and other scene components on established urban driving benchmarks (Cordts et al., 2016). More recently, vision foundation models have further expanded the scope of automated image segmentation. The Segment Anything Model (SAM; Kirillov et al., 2023), trained on a dataset containing more than one billion segmentation masks, enables class-agnostic object segmentation through visual prompts, including point inputs. Its computationally efficient derivative, MobileSAM (Zhang et al., 2023), extends prompt-based segmentation to resource-constrained computing environments, including central processing unit (CPU)-based workstations. These developments have made foundation-model-assisted annotation increasingly accessible beyond computationally intensive research environments, provided that suitable annotation platforms are available to integrate such models into practical data-labelling workflows.

The performance of any supervised semantic segmentation model is fundamentally dependent on the quality and consistency of its training annotations. Generating polygon-level labels for road-scene imagery is particularly demanding because road environments contain complex boundaries, irregular geometric configurations, occlusions, and spatially adjacent features. For instance, the Cityscapes benchmark required approximately 1.5 hours of expert annotation effort per image (Cordts et al., 2016), while manual vertex placement can introduce inter-annotator variability in boundary delineation, potentially compromising ground-truth consistency (Wang et al., 2025). Consequently, manual annotation becomes a substantial constraint when developing large-scale, high-resolution datasets. Interactive annotation approaches, such as Polygon-RNN (Castrejon et al., 2017) and Polygon-RNN++ (Acuna et al., 2018), address this limitation by predicting polygon vertices sequentially, thereby reducing the amount of manual input required for object delineation. Similarly, prompt-based segmentation approaches, including SAM, can generate object masks from sparse visual inputs, substantially reducing manual boundary-tracing requirements. Beyond interactive segmentation, model-assisted annotation introduces a progressive learning paradigm in which models trained on an initial set of labelled images generate segmentation proposals for subsequent samples. This shifts the annotator's primary responsibility from manual delineation towards reviewing and correcting model-generated predictions, establishing a human-in-the-loop annotation process that can reduce labelling effort while promoting annotation consistency across successive annotation sessions (Hsu, 2012).

Despite the increasing maturity of open-source annotation platforms, including LabelMe (Russell et al., 2008), the VGG Image Annotator (Dutta & Zisserman, 2019), and the Computer Vision Annotation Tool (CVAT), as well as commercial solutions such as Roboflow, several limitations remain relevant to road safety research. First, many open-source annotation platforms provide limited native integration of artificial intelligence (AI)-assisted segmentation, while access to foundation models such as SAM may depend on external plugins, additional configuration, or cloud-based services. Second, the pretrained segmentation models commonly available through existing platforms predominantly address general-purpose object categories and may not adequately represent specialized road infrastructure features, such as pavement surfaces, edge lines, rumble strips, and roadside furniture, which are not explicitly represented as dedicated categories in widely used datasets such as the Common Objects in Context (COCO) dataset (Lin et al., 2014) and the Pascal Visual Object Classes (VOC) dataset (Everingham et al., 2010). Third, existing open-source annotation workflows provide limited support for training lightweight, task-specific segmentation models using an incrementally expanding collection of user-generated annotations and deploying the updated models for annotation assistance within the same application session. Fourth, general-purpose annotation platforms do not necessarily incorporate road-scene-specific annotation conventions, class definitions, or workflow guidance required for systematically labelling transportation infrastructure and roadside features. These limitations motivate the development of an integrated annotation environment that combines conventional image-processing methods, foundation-model-based segmentation, pretrained semantic models, and user-trained segmentation networks.

To address these limitations, this paper introduces RoadAnnoter, a self-contained, open-source desktop application developed for polygon-based annotation of road-scene imagery. Implemented as a unified Python application using the PyQt5 framework, RoadAnnoter integrates six complementary annotation modes: (i) manual polygon delineation with undo/redo functionality and multi-class annotation support; (ii) SAM Prompt, which employs point-based visual prompts with SAM or MobileSAM (Kirillov et al., 2023; Zhang et al., 2023); (iii) Smart Selection, an interactive SAM-based segmentation mode incorporating an adaptive bounding-box prior to facilitate single-click object selection; (iv) Edge Fill, a conventional image-processing approach combining multi-scale Canny edge detection with colour-similarity-based region filling in the CIELAB colour space, without requiring a deep learning model; (v) pretrained semantic segmentation suggestions generated using SegFormer models fine-tuned on the Cityscapes dataset (Xie et al., 2021; Cordts et al., 2016), providing predictions for relevant road-scene semantic categories; and (vi) task-specific segmentation suggestions generated using TinySegNet, a lightweight neural network based on the MobileNetV3-Small architecture (Howard et al., 2019), which can be incrementally trained using annotations generated within the application. The six modes are integrated through a common mask-to-polygon conversion pipeline that applies post-processing operations to generate simplified polygon boundaries and exports the resulting annotations in the You Only Look Once (YOLO) segmentation format. RoadAnnoter was developed and deployed for annotating street-view images acquired at recorded road accident locations in Bangladesh, supporting a vision-language model-based investigation of free-flow speed estimation.

The remainder of this paper is organized as follows. Section 2 reviews the relevant literature on image-processing-based annotation assistance, deep learning-based semantic segmentation, and polygon generation. Section 3 presents the methodological framework underlying the proposed annotation approaches. Section 4 describes the software architecture, implementation, and graphical user interface of RoadAnnoter. Section 5 introduces the dataset used in the study. Section 6 demonstrates the application of the six annotation modes through representative road safety annotation scenarios. Finally, Section 7 presents the conclusions and recommendations for future development.

## 2 Literature Review

Road-scene annotation assistance draws upon three complementary technical domains: classical image processing for region delineation, deep learning-based segmentation for automated feature recognition, and computational geometry algorithms for converting pixel-level segmentation masks into polygonal representations. These techniques provide different levels of automation, computational complexity, and dependence on training data. This section reviews the theoretical foundations and methodological developments within each domain, emphasizing their capabilities and limitations in the context of road and roadside feature annotation.

### 2.1 Image Processing as Annotation Assistance

#### 2.1.1 Edge Detection Methods

Edge detection identifies spatial discontinuities in image intensity that frequently correspond to object boundaries and transitions between adjacent regions. Early gradient-based operators, including the Roberts, Sobel, and Prewitt operators, estimate first-order intensity derivatives using small convolution kernels. Although computationally efficient, these operators are sensitive to image noise and may produce thick or fragmented edge responses, limiting their effectiveness in complex outdoor environments. The Canny edge detector (Canny, 1986) provides a more systematic approach by incorporating Gaussian smoothing, gradient estimation, non-maximum suppression, and dual-threshold hysteresis. The smoothing operation reduces high-frequency noise, while non-maximum suppression localizes edge responses by retaining local gradient maxima. Subsequently, hysteresis thresholding uses upper and lower thresholds to distinguish strong edges from weak responses, retaining weak edges only when they are connected to sufficiently strong edge segments. The resulting detector addresses three principal criteria: detection accuracy, localization precision, and minimal multiple responses to a single edge. These properties make Canny particularly relevant to road-scene imagery, where pavement markings, shadows, surface deterioration, and variations in illumination frequently produce complex and low-contrast boundaries.

In interactive annotation workflows, detected edges can function as spatial barriers that constrain region-growing and flood-filling operations. However, texture variations within relatively homogeneous surfaces, such as asphalt pavement, may generate internal edge responses that interrupt otherwise continuous regions. To mitigate this limitation, edge detection can be preceded by bilateral filtering (Tomasi & Manduchi, 1998), which suppresses local intensity variations while preserving prominent intensity discontinuities through joint spatial and photometric weighting. Furthermore, multi-scale Canny detection can combine edge responses obtained using different threshold configurations to capture boundaries across varying contrast levels, including fine pavement markings and broader transitions between road surfaces and adjacent environments. Nevertheless, edge-based annotation remains sensitive to threshold selection, local image texture, and illumination conditions. Highly textured surfaces may generate fragmented edge structures, whereas weak or discontinuous boundaries can result in leakage during subsequent region-filling operations. Additionally, seed points located directly on detected edges may produce ambiguous region initialization, requiring appropriate seed placement or preprocessing.

### 2.1.2 Colorimetric Segmentation

Colour-based segmentation partitions an image into spatial regions according to the similarity of pixel-level colour characteristics. This approach is particularly useful for road-scene annotation because several environmental features, including pavement surfaces, lane markings, vegetation, and sidewalks, may exhibit locally consistent colour distributions. Seeded flood filling is a computationally efficient region-selection technique in which neighbouring pixels are progressively included when their colour difference from a reference seed or an evolving region representative satisfies a predefined similarity criterion. Under a Euclidean colour-distance formulation, a pixel is accepted when its colour-space distance from the reference colour remains within a specified tolerance. The CIELAB colour space is frequently preferred over conventional RGB representation because it provides a perceptually oriented colour coordinate system in which Euclidean distances approximate perceived colour differences more consistently. Consequently, colour-distance thresholds in CIELAB space can provide a more meaningful basis for region selection under varying illumination and surface appearance.

Several more sophisticated approaches have been developed to improve colour- and region-based image segmentation. Mean-shift segmentation (Comaniciu & Meer, 2002) identifies high-density regions in a joint spatial and colour feature space through iterative density-mode seeking, enabling the identification of locally homogeneous image regions. Watershed segmentation (Vincent & Soille, 1991) interprets an image gradient as a topographic surface and simulates flooding from regional minima to establish boundaries between adjacent catchment basins. GrabCut (Rother et al., 2004) formulates foreground-background separation as an iterative optimization problem involving colour distribution modelling and graph-cut-based energy minimization. Similarly, Simple Linear Iterative Clustering (SLIC) superpixels (Achanta et al., 2012) groups neighbouring pixels into compact, approximately uniform regions by combining spatial proximity and colour similarity. Although these methods offer more structured segmentation than basic flood filling, their effectiveness depends substantially on the underlying assumptions regarding regional homogeneity and boundary contrast. Such assumptions may be violated in road environments characterized by weathered asphalt, cast shadows, heterogeneous vegetation, surface contamination, and wet pavement. Therefore, colour-based segmentation is particularly suitable for interactively selecting locally homogeneous regions but may require additional boundary constraints or model-based assistance when applied to visually heterogeneous road-scene features.

### 2.1.3 Morphological Operations

Mathematical morphology provides a set of nonlinear image-processing operations based on the interaction between binary or grayscale images and predefined structuring elements (Serra, 1982). The fundamental morphological operations include dilation and erosion, from which opening and closing are derived. Dilation expands foreground regions according to the geometry of the structuring element, whereas erosion contracts them by removing boundary pixels that do not satisfy the corresponding neighbourhood condition. Morphological opening, defined as erosion followed by dilation, suppresses small isolated foreground regions and narrow protrusions. Conversely, morphological closing, defined as dilation followed by erosion, fills small internal gaps and bridges narrow discontinuities between adjacent foreground regions. In annotation workflows, these operations can be applied to binary segmentation masks generated through manual interaction, conventional image processing, or deep learning inference to remove localized noise, improve region connectivity, and regularize boundary geometry. The resulting masks provide cleaner inputs for subsequent contour extraction and polygon generation.

Following mask refinement, polygon simplification reduces the number of vertices required to represent the segmented boundary. The Douglas–Peucker algorithm (Douglas & Peucker, 1973) is a widely adopted geometric simplification technique that recursively approximates a polyline using a reduced set of vertices. Starting with the endpoints of a contour segment, the algorithm identifies the intermediate vertex with the maximum perpendicular distance from the corresponding line segment. If this distance exceeds a predefined tolerance, the contour is divided at that vertex and the procedure is recursively repeated for the resulting subsegments. Otherwise, the intermediate vertices are discarded. By defining the tolerance relative to the original contour perimeter, the simplification process can be adapted to the spatial scale of different objects. This approach substantially reduces polygon complexity while retaining the principal geometric characteristics of the original segmentation boundary, thereby facilitating efficient annotation storage and downstream model training.

## 2.2 Deep Learning Models as Annotation Assistance

### 2.2.1 Pixel-Level Semantic Segmenters

Fully convolutional networks (FCNs) introduced an end-to-end framework for dense pixel-level prediction by replacing conventional fully connected classification layers with convolutional operations and spatial upsampling mechanisms (Long et al., 2015). This architectural development established a foundation for subsequent semantic segmentation models capable of assigning predefined semantic categories to individual image pixels. DeepLabV3+ (Chen et al., 2018) further advanced this framework through atrous spatial pyramid pooling (ASPP), which employs parallel atrous convolutions at multiple dilation rates to capture contextual information across different spatial scales. Its encoder-decoder architecture additionally combines high-level semantic representations with lower-level spatial information to improve boundary localization. These architectures have been extensively investigated using benchmark datasets such as COCO (Lin et al., 2014) and Pascal Visual Object Classes (VOC) (Everingham et al., 2010). However, the standard category taxonomies of these datasets provide limited representation of specialized road infrastructure features, restricting the direct applicability of models trained exclusively on their standard annotations to detailed road-scene characterization.

The Cityscapes dataset (Cordts et al., 2016) provides a more specialized benchmark for urban scene understanding, comprising 5,000 finely annotated images and 19 semantic classes covering road surfaces, sidewalks, buildings, vegetation, sky, pedestrians, vehicles, and other urban infrastructure components. Its annotation taxonomy is consequently more closely aligned with the requirements of road-scene segmentation than general-purpose object recognition datasets. Among the architectures developed for semantic segmentation, SegFormer (Xie et al., 2021) combines a hierarchical Mix Transformer encoder with a lightweight multilayer perceptron (MLP) decoder to generate multiscale semantic representations without relying on conventional positional encoding or complex decoder structures. Its architecture provides an effective balance between contextual feature extraction and computational efficiency. SegFormer models fine-tuned on Cityscapes therefore offer a relevant source of pretrained semantic predictions for road and roadside feature annotation, particularly for classes represented within the Cityscapes taxonomy. Nevertheless, their applicability remains constrained by the dataset's predefined class distribution and annotation conventions, making additional annotation approaches necessary for identifying specialized road safety features not explicitly represented in the pretrained model outputs.

The Segment Anything Model (SAM; Kirillov et al., 2023) introduces a fundamentally different paradigm based on promptable, class-agnostic image segmentation. Trained using the Segment Anything 1 Billion (SA-1B) dataset, comprising approximately 11 million images and more than one billion segmentation masks, SAM is designed to generate object masks from user-provided prompts, including foreground and background points, bounding boxes, and existing mask inputs. Unlike conventional semantic segmentation networks, which assign pixels to a predefined set of semantic categories, SAM identifies spatial regions corresponding to the supplied prompts without requiring a predefined task-specific class taxonomy. This prompt-based formulation is particularly advantageous for interactive annotation because an annotator can specify a region of interest through a small number of visual inputs and obtain a corresponding segmentation mask for subsequent verification and refinement. MobileSAM (Zhang et al., 2023) extends this capability to resource-constrained environments by replacing SAM's computationally intensive image encoder with a lightweight Tiny Vision Transformer (TinyViT)-based encoder (Wu et al., 2022), while retaining the prompt encoder and mask decoder framework. This architectural adaptation reduces computational requirements and facilitates the deployment of prompt-based segmentation on relatively modest hardware, making interactive annotation more accessible in environments where dedicated graphics processing units (GPUs) or cloud-based inference services are unavailable.

### 2.2.2 Lightweight Architectures for Incremental Model-Assisted Annotation

Model-assisted annotation introduces an additional requirement beyond conventional segmentation inference: the underlying model must be trainable and deployable using relatively small collections of labelled images and limited computational resources. Unlike large-scale benchmark training, which typically involves extensive datasets and substantial computational infrastructure, an annotation-oriented model should support rapid adaptation to domain-specific features as new labelled samples become available. MobileNetV3-Small (Howard et al., 2019) provides a suitable architectural foundation for this purpose. Developed through hardware-aware neural architecture search, MobileNetV3 incorporates computationally efficient building blocks, including depthwise separable convolutions, squeeze-and-excitation modules, and hard-swish activation functions. Its compact architecture reduces the computational and parameter requirements associated with feature extraction compared with conventional high-capacity convolutional networks and transformer-based encoders.

When combined with a lightweight decoder, such as an atrous spatial pyramid pooling module and a final $1 \times 1$ convolutional classification layer, a MobileNetV3-Small-based segmentation network can provide a computationally efficient framework for learning pixel-level semantic representations. The resulting model can be adapted using annotator-generated labels to produce class-specific segmentation predictions for subsequent images. Within a human-in-the-loop annotation workflow, these predictions serve as preliminary polygon suggestions that can be reviewed, corrected, and accepted by the annotator rather than requiring complete manual delineation. The accepted annotations can subsequently be incorporated into the available training dataset to support further model updates. This progressive interaction between annotation generation, model adaptation, and prediction verification establishes an iterative learning process through which task-specific segmentation capability can evolve with the accumulation of labelled samples. The effectiveness of this approach, however, depends on the representativeness and consistency of the accumulated annotations, as well as the training strategy employed during successive model updates.

### 2.3 Connected Component Algorithms for Polygon Formation

Segmentation methods based on deep learning, flood filling, and other image-processing operations generally produce pixel-level binary masks rather than the ordered vertex sequences required for polygon-based annotation formats. Converting these masks into polygon representations requires the identification of spatially connected foreground regions, extraction of their boundaries, and geometric simplification. Connected-component labelling provides the initial operation by assigning a unique integer label to each maximal connected foreground region according to a predefined neighbourhood connectivity criterion. Under eight-connectivity, each foreground pixel is considered connected to its horizontal, vertical, and diagonal neighbours. Consequently, spatially disconnected regions are identified as separate components, allowing individual candidate objects to be extracted from a single binary mask. A minimum component-area threshold can subsequently be applied to eliminate isolated noise regions and other insignificant foreground components before polygon generation.

Following component identification, contour extraction determines the ordered boundary coordinates of each retained region. The Suzuki–Abe border-following algorithm (Suzuki & Abe, 1985), implemented in OpenCV through the findContours function, provides an efficient method for tracing the external and internal boundaries of binary image regions while preserving their contour hierarchy. For polygon-based annotation, the external contour of each component can be extracted to represent its outer geometric boundary, while internal contours may be retained when the annotation requirements necessitate the representation of holes. Before contour extraction, Gaussian smoothing followed by binary re-thresholding can be applied to reduce pixel-level irregularities and staircase artefacts along the mask boundary. The resulting contour is then approximated using the Douglas–Peucker algorithm (Douglas & Peucker, 1973), which reduces the number of boundary vertices while maintaining the principal geometric structure of the segmented region. The simplification tolerance controls the trade-off between polygon complexity and boundary fidelity, and contours containing fewer than three distinct vertices are discarded because they cannot define a valid polygon. The resulting ordered polygon coordinates can subsequently be normalized with respect to image width and height and stored in the YOLO segmentation format. By applying a common mask-to-polygon conversion procedure across all annotation modes, the proposed framework maintains a consistent geometric representation and annotation output structure, irrespective of the segmentation technique used to generate the initial binary mask.

## 3 Methodology

### 3.1 Annotation Framework Design

RoadAnnoter (**Figure 1**) is designed around a human-in-the-loop annotation paradigm in which automated and semi-automated segmentation methods generate candidate polygon annotations that require explicit human verification before being incorporated into the project dataset. This design ensures that the annotator retains final authority over all generated annotations and prevents unverified model predictions from being committed to the annotation files. Each assisted annotation mode presents its segmentation proposal as a dashed cyan overlay on the image canvas, allowing the annotator to inspect the predicted boundary before accepting or rejecting it. A proposal is committed to the annotation dataset only following an explicit acceptance action, which can be performed using the Enter or Space key or by double-clicking the proposed region. Conversely, the proposal can be discarded using the Escape key or a right-click action. Manual polygon delineation follows a separate interaction procedure in which the annotator directly specifies the polygon vertices by clicking successive boundary locations on the image canvas. The polygon is completed by pressing Enter or double-clicking the initial vertex. This separation between assisted mask-based annotation and direct polygon construction provides a consistent interaction framework while accommodating different levels of automation and user control.

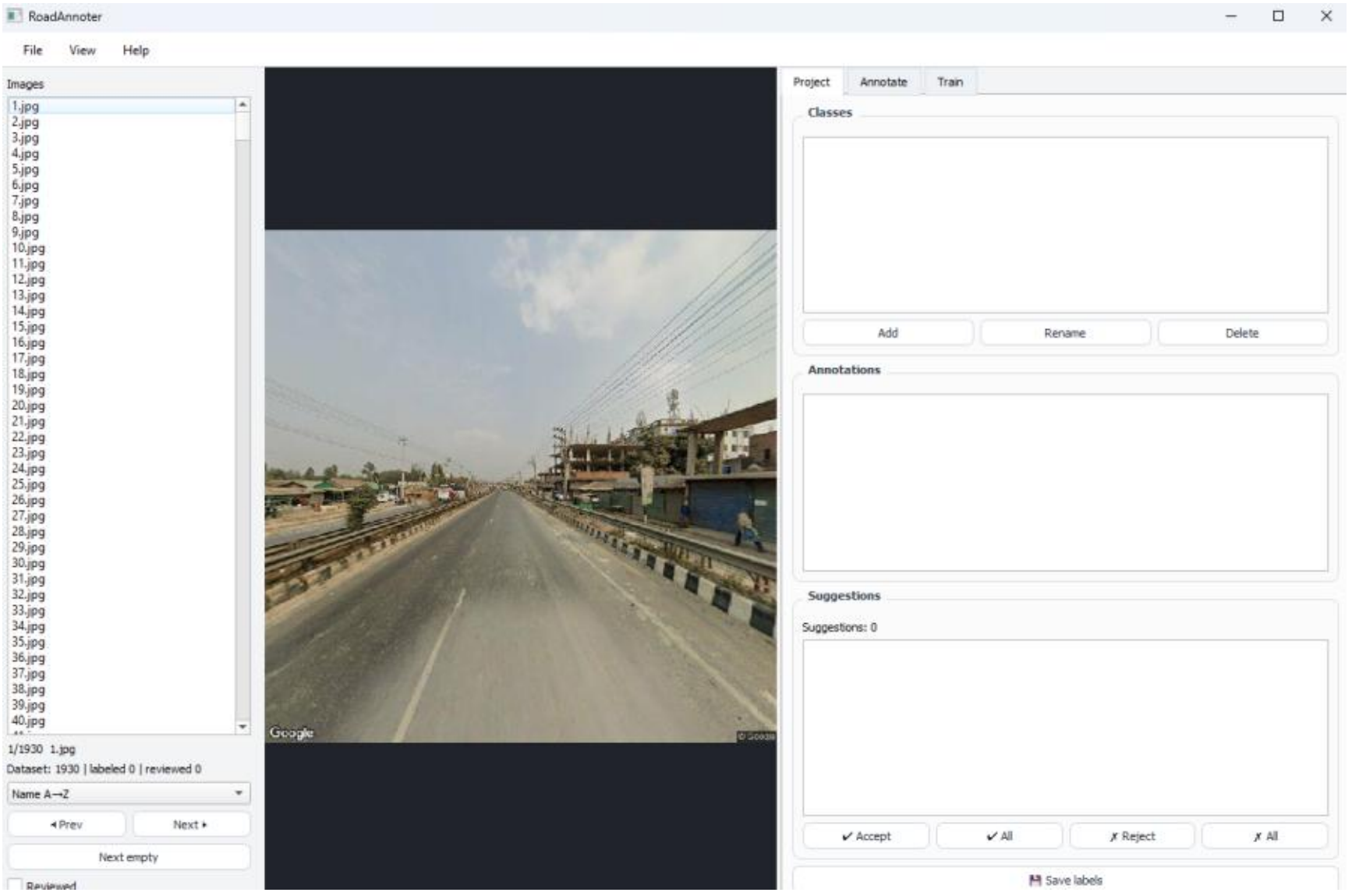

**Figure 1.** RoadAnnoter Graphical User Interface (GUI)

The six annotation modes are organized according to their computational complexity, model requirements, and dependence on external infrastructure. Manual polygon delineation requires only the core PyQt5 and OpenCV installation, without additional model weights or external connectivity. Edge Fill similarly operates using conventional image-processing operations implemented through OpenCV and can be executed on central processing unit (CPU)-based workstations without an internet connection. SAM Prompt and Smart Selection require a locally available SAM or MobileSAM model checkpoint, with storage requirements ranging approximately from 40 to 375 MB depending on the selected model variant; however, subsequent annotation operations do not require internet connectivity. Pretrained semantic segmentation suggestions require an initial internet connection to retrieve the corresponding model weights from the Hugging Face model hub and the PyTorch model repository. Once downloaded and cached locally, these models can be loaded and executed in subsequent sessions without external connectivity. Own-model suggestions become available following the completion of an initial training session, with model availability determined by the successful recording of at least one training epoch. This graduated dependency architecture enables RoadAnnoter to support annotation activities across a range of computational environments, from standalone CPU-based workstations to GPU-equipped research systems, without making cloud connectivity a prerequisite for routine annotation operations.

The project architecture employs a file-based annotation management system in which each image is associated with a corresponding set of project metadata and annotation records. Specifically, the annotation structure comprises up to three file types: (i) an image-specific text file (.txt), stored alongside the corresponding image and sharing its base filename, containing polygon annotations in the YOLO segmentation format; (ii) a project-level classes.txt file containing the ordered list of semantic class names and their corresponding numerical indices; and (iii) an optional reviewed-marker file, implemented as a zero-byte file, indicating that the annotator has inspected the corresponding image and designated it as complete. This structure maintains a direct association between image data, annotation records, and project-level class definitions while avoiding the need for a separate database management system. To improve data integrity during annotation updates, label files are written using an atomic file-replacement procedure. Specifically, updated annotation content is first written to a temporary file, which subsequently replaces the original label file through the operating system's os.replace operation. This approach minimizes the risk of incomplete annotation records resulting from interruptions during file-writing operations. Furthermore, the image canvas maintains a snapshot-based undo/redo mechanism with a maximum history depth of 100 states. The history records modifications associated with vertex repositioning, polygon insertion and deletion, class reassignment, and batch acceptance of segmentation suggestions. Consequently, annotators can reverse or restore successive editing operations while maintaining the consistency of the annotation state.

Semantic class management is implemented through a centralized class registry maintained in the project-level classes.txt file. Each annotation is stored using an integer-valued class index rather than the corresponding textual class name, with the class index resolved against the current registry whenever annotations are displayed or exported. This separation between annotation identifiers and class definitions facilitates project-level class management without requiring the modification of individual polygon coordinates when a class name is changed. When a class is deleted, all annotations associated with its index are removed, and the indices of subsequent classes are adjusted to maintain a contiguous class-index sequence. Corresponding annotation records are updated to preserve consistency between the stored labels and the revised class registry. The framework supports multiple semantic classes within an individual image, with the active annotation class selected interactively through the class-management panel or the designated keyboard shortcut (L). This centralized approach provides a consistent mechanism for managing class assignments across images and maintaining compatibility with downstream segmentation model training workflows.

## 3.2 Edge Fill: Classical Edge-Guided Flood Fill

Edge Fill is designed for annotation environments in which deep learning models are unavailable, unnecessary, or computationally impractical. It is particularly suitable for CPU-only workstations, air-gapped research environments, and the initial stages of an annotation project before a Segment Anything Model (SAM) checkpoint has been obtained. The method combines two complementary sources of image information: structural boundary information derived from edge detection and regional appearance information obtained through colour-similarity-based segmentation. Edge detection identifies potential object boundaries, whereas colour-based region selection identifies pixels exhibiting similar chromatic characteristics to a user-selected seed location. Neither mechanism alone is sufficiently reliable for heterogeneous road-scene imagery. Edge maps may contain fragmented or excessive responses in textured asphalt regions, while unconstrained colour-based region growing may extend beyond the intended object at boundaries characterized by weak intensity or colour contrasts. Edge Fill addresses these limitations by restricting region propagation to pixels that satisfy both a predefined colour-similarity criterion and the spatial constraints imposed by detected edge barriers. The complete procedure comprises six sequential processing stages.

**Step 1: Bilateral smoothing.** The input image is first converted into grayscale and subsequently smoothed using a bilateral filter with a neighbourhood diameter of d = 9 pixels and spatial and intensity-domain standard deviations of $\sigma_s = 75$ and $\sigma_r = 75$, respectively (Tomasi & Manduchi, 1998). Unlike conventional Gaussian smoothing, which determines pixel contributions primarily through spatial proximity, bilateral filtering considers both spatial distance and photometric similarity. Consequently, pixels that are spatially distant or exhibit substantially different intensity values receive lower weights during the filtering process. The filtered intensity at a pixel location p is expressed as:

$$I_f(p) = (1/W_p) \times \Sigma_{\{q \in \Omega_p\}} \, exp\left[-\frac{\left\|p-q\right\|^2}{2{\sigma_s}^2}\right] \times exp\left[-\frac{\left(I(p)-I(q)\right)^2}{2{\sigma_r}^2}\right] \times I(q) \quad (1)$$

where $I(p)$ and $I(q)$ denote the original grayscale intensities at locations $p$ and $q$, respectively; $\Omega_p$ represents the filtering neighbourhood; $\sigma_s$ controls spatial weighting; $\sigma_r$ controls intensity-based weighting; and $W_p$ is the normalization factor ensuring that the filter weights sum to unity. The spatial component attenuates contributions from distant pixels, whereas the intensity-domain component reduces the influence of neighbouring pixels with substantially different grayscale values. This combination suppresses local photometric variations while preserving sufficiently prominent intensity transitions. It therefore reduces some of the unnecessary internal edge responses associated with pavement texture before subsequent edge detection.

**Step 2: Multi-threshold Canny edge detection.** The bilateral-filtered grayscale image is processed using the Canny edge detector (Canny, 1986) with two different pairs of lower and upper hysteresis thresholds. The first edge map is generated using the user-configurable threshold values, with default lower and upper thresholds of 20 and 80, respectively. A second edge map is generated using half these threshold values, corresponding to lower and upper thresholds of 10 and 40. The resulting binary edge maps are combined through a pixel-wise bitwise OR operation:

$$E_c(x,y) = E_1(x,y) \mid E_2(x,y) \quad (2)$$

where $E_1(x,y)$ and $E_2(x,y)$ represent the edge responses obtained from the two threshold configurations, and $E_c(x,y)$ denotes the combined edge map. The use of two threshold configurations provides complementary edge responses by allowing weaker gradient transitions to be retained alongside more prominent boundaries. The lower-threshold configuration generally increases sensitivity to weak and low-contrast transitions, whereas the higher-threshold configuration provides a more selective response to stronger intensity gradients. Their combination increases the availability of candidate boundary information across road-scene features with varying contrast characteristics without requiring the annotator to adjust separate threshold pairs for individual boundary types. This procedure constitutes multi-threshold edge detection rather than multi-scale image analysis because both edge maps are generated from the same spatially filtered image.

**Step 3: Edge barrier construction.** The combined edge map is subsequently dilated to generate spatial barrier regions that constrain the flood-filling operation. Dilation is performed twice using an elliptical structuring element with dimensions of $(2r + 1) \times (2r + 1)$, where the default radius is $r = 4$ pixels, resulting in a 9 × 9 kernel. The resulting barrier map is represented as:

$$B(x,y) = (E_c \oplus K_r) \oplus K_r \quad (3)$$

where $\oplus$ denotes morphological dilation and $K_{\mathrm{r}}$ represents the elliptical structuring element. The two dilation iterations expand the detected edge regions, increasing their spatial continuity and reducing the likelihood of region propagation through small discontinuities in the original edge map. The elliptical kernel provides a more isotropic expansion than a square structuring element, reducing directional bias in the resulting barrier geometry. This is particularly relevant to road-scene boundaries characterized by curved or irregular geometries, including curb lines, pavement edges, and lane-marking boundaries. The resulting barrier map provides a spatial constraint for the colour-based region selection performed in the subsequent stage.

**Step 4: Colour-similarity mapping.** The original input image is transformed from the red-green-blue (RGB) colour space into the CIELAB colour space. The colour vector of the user-selected seed pixel is represented as $c_s = [L_s *, a_s *, b_s *]^{\mathrm{T}}$, while the colour vector of an arbitrary image pixel $(x,y)$ is represented as $c(x,y) = [L*(x,y), a*(x,y), b*(x,y)]^{\mathrm{T}}$. The colour difference between the seed pixel and each image pixel is calculated using the Euclidean distance in CIELAB space:

$$\Delta E_{ab(x,y)} = \sqrt{\{[L(x,y) - L_s]^2 + [a*(x,y) - a_s*]^2 + [b*(x,y) - b_s*]^2\}} \quad (4)$$

A binary colour-similarity map is subsequently generated using a predefined tolerance $\tau_c$:

$$C(x,y) = 1, if\ \Delta E_{ab}(x,y) \le \tau_c;\ otherwise, C(x,y) = 0 \quad (5)$$

The default tolerance is $\tau_c = 20$ CIELAB units. CIELAB is preferred over conventional RGB coordinates because its approximately perceptually uniform characteristics provide a more interpretable representation of colour differences. In particular, the separation of luminance from chromaticity facilitates colour-based region selection when neighbouring pixels exhibit variations in brightness and colour composition. However, the selected tolerance represents an operational segmentation parameter rather than a universal perceptual discrimination threshold. The effectiveness of this parameter depends on the local appearance of the target feature, illumination conditions, and the degree of colour variation within the region of interest. Accordingly, the tolerance can be adjusted by the annotator to accommodate different road-surface conditions and roadside feature characteristics.

**Step 5: Barrier-constrained flood filling.** The colour-similarity map and the edge barrier map are combined to establish the set of pixels eligible for region propagation. Specifically, the reachability map is obtained through a pixel-wise logical conjunction between the colour-similarity map and the complement of the edge barrier map:

$$R(x,y) = C(x,y)\ AND\ NOT\ B(x,y) \quad (6)$$

where $R(x,y)$ identifies pixels that simultaneously satisfy the colour-similarity criterion and are not occupied by detected edge barriers. The resulting binary map therefore restricts flood propagation to spatially connected regions that exhibit sufficient colour similarity to the selected seed pixel. OpenCV's `floodFill` function is subsequently employed in *FLOODFILL_MASK_ONLY* mode, using zero lower and upper intensity-difference tolerances (loDiff = upDiff = 0). This configuration enables flood filling to operate on the preconstructed binary eligibility map rather than independently evaluating colour differences between neighbouring pixels during propagation. The seed coordinates are offset by one pixel in both image dimensions to accommodate the additional border required by the OpenCV flood-fill mask convention. The resulting connected region represents the portion of the colour-similar area that is spatially accessible from the selected seed without crossing the constructed edge barriers. To eliminate small isolated regions and insignificant segmentation artefacts, connected regions containing fewer than 500 pixels are discarded before morphological refinement.

**Step 6: Morphological refinement.** The flood-filled binary mask is subjected to successive morphological operations to improve region continuity and suppress residual boundary irregularities. First, morphological closing is performed using an 11 × 11 elliptical structuring element. Closing, defined as dilation followed by erosion, bridges narrow discontinuities and fills sufficiently small internal gaps within the segmented region. This operation helps recover locally incomplete regions arising from colour variations or thin edge responses that may have excluded otherwise relevant pixels during flood filling. Subsequently, morphological opening is applied using a 5 × 5 elliptical structuring element. Opening, defined as erosion followed by dilation, suppresses small isolated foreground components and narrow protrusions that may result from unintended region propagation through discontinuities in the edge barriers. The combination of these operations produces a more spatially coherent binary mask with reduced local irregularities. The refined mask is then forwarded to the common mask-to-polygon conversion pipeline described in Section 3.6, where contour extraction, polygon simplification, and coordinate normalization are performed to generate the final annotation in the YOLO segmentation format.

## 3.3 SAM-Based Annotation: SAM Prompt and Smart Selection

RoadAnnoter integrates two variants of the Segment Anything Model (SAM): the original SAM architecture (Kirillov et al., 2023) and MobileSAM (Zhang et al., 2023). These models provide prompt-guided segmentation capabilities through point and bounding-box interactions, allowing annotators to generate object masks without manually tracing polygon boundaries. The application supports automatic checkpoint-type detection by inspecting the keys contained in the loaded model state dictionary. In the current implementation, the presence of relative-position embedding keys, such as *`attn.rel_pos_h`*, is used as an indicator of an original SAM checkpoint, whereas their absence is used to identify MobileSAM checkpoints. When the official *`mobile_sam`* Python package is installed, MobileSAM checkpoints are loaded through its corresponding model implementation. This is the recommended loading approach because it ensures that the model architecture, including the TinyViT image encoder, is instantiated consistently with the checkpoint configuration.

For both model variants, the input image is resized and padded to a square resolution of 1024 × 1024 pixels before being processed by the image encoder. The encoder generates a spatial feature embedding with 256 channels and a spatial resolution of 64 × 64 pixels. This image embedding is computed once when an image is loaded and subsequently cached for reuse during interactive annotation. Consequently, additional point or bounding-box prompts do not require the image encoder to process the entire image again, reducing the computational cost of successive segmentation requests. The cached embedding is passed to the prompt encoder and mask decoder whenever a new prompt configuration is submitted. The model is configured to use a CUDA-enabled graphics processing unit (GPU) when available and otherwise falls back to central processing unit (CPU) execution. Under the tested configurations, prompt evaluation typically takes less than 50 ms per interaction with the original SAM ViT-B encoder on a GPU and less than 100 ms on a CPU with MobileSAM. These execution times should be interpreted as implementation-specific observations rather than hardware-independent performance guarantees.

The SAM prompt encoder transforms user-provided point coordinates, point labels, bounding boxes, and optional coarse mask inputs into sparse and dense embeddings in the model's 256-dimensional feature space. These prompt representations are combined with the cached image embedding and processed by the lightweight transformer-based mask decoder. With multimask prediction enabled, the decoder generates three candidate segmentation masks, each accompanied by a predicted intersection-over-union (IoU) quality score. RoadAnnoter implements two distinct interaction strategies using this common prediction architecture: SAM Prompt, which prioritizes the model's predicted mask quality, and Smart Selection, which prioritizes the spatial coverage of the segmented region.

**SAM Prompt.** SAM Prompt provides an iterative point-based segmentation interface in which the annotator progressively specifies the foreground and background regions of the target object. Each left-click adds a foreground point prompt at the selected image coordinate, whereas each right-click adds a background exclusion point. The accumulated point coordinates and their corresponding binary labels (1 = foreground and 0 = background) are passed to the `SamPredictor.predict` function with *`multimask_output = True`*. The model consequently returns three candidate masks together with their predicted IoU scores. RoadAnnoter selects the candidate with the highest predicted IoU and displays it as a pending annotation suggestion, replacing any previously generated suggestion within the current interaction session.

The annotator can iteratively add foreground or background points to refine the predicted segmentation before accepting the result. Foreground prompts guide the model towards the intended object region, while background prompts help exclude adjacent regions that may otherwise be incorporated into the prediction. This capability is particularly useful in road-scene imagery, where neighbouring features such as curbs, drainage covers, and road markings may exhibit visual characteristics similar to those of the target surface. Once the proposed mask is accepted, the accumulated prompt buffer is cleared to initialize the next annotation interaction. This iterative interaction strategy allows the annotator to progressively resolve segmentation ambiguities while retaining the model's predicted-IoU-based candidate selection.

**Smart Selection.** Smart Selection is designed as a simplified, single-click interaction mode for segmenting relatively large and spatially continuous road and roadside features. A single foreground click triggers a combined point-and-bounding-box prompt. The bounding box is centred on the clicked image coordinate and extends in both directions by a configurable fraction of the image width and height. The default Smart Coverage setting is 40%, with an adjustable range of 10%–100%. Thus, the selected coverage parameter controls the spatial extent of the bounding box relative to the image dimensions, with the actual box constrained by the image boundaries.

The combined point-and-box prompt provides the model with two complementary forms of spatial information. The foreground point identifies an interior location associated with the target object, while the bounding box provides a broader spatial constraint on the expected segmentation extent. This combination is intended to improve the prediction of complete, spatially extensive features, such as road carriageways and contiguous roadside vegetation, for which a point-only prompt may produce a spatially restricted mask. In particular, the bounding box supplies contextual information about the approximate extent of the target region and can help reduce incomplete segmentation of large, relatively homogeneous surfaces.

Unlike SAM Prompt, which selects the candidate mask with the highest predicted IoU, Smart Selection selects the candidate with the largest segmented area among the three returned masks. This selection strategy is intended to favour spatial completeness when annotating extensive road surfaces and other large roadside features, for which incomplete coverage may be more consequential than minor boundary imprecision. However, selecting the largest mask does not necessarily identify the candidate with the highest boundary accuracy or predicted IoU; it represents a deliberate implementation-level preference for spatial coverage. Each Smart Selection interaction independently initializes the point-and-box prompt and resets the prompt buffer. Consequently, successive clicks generate independent segmentation suggestions without accumulating foreground or background prompts from previous objects.

**Optional edge snapping (experimental; disabled by default).** RoadAnnoter also incorporates an optional edge-snapping procedure for refining the polygon representation of SAM-generated masks. When enabled, each contour vertex extracted from the predicted mask is examined for a nearby Canny edge response within a configurable search radius, with a default radius of 4 pixels. The vertex may be relocated to the nearest eligible edge pixel, subject to the constraint that the candidate snapping location must remain within the original SAM mask boundary. This restriction prevents the refinement procedure from displacing polygon vertices beyond the spatial extent of the original segmentation mask. If no eligible edge pixel is identified within the specified search neighbourhood, the vertex retains its original position.

The edge map used for this operation is generated independently of the Edge Fill procedure, using separate Canny thresholds with default lower and upper values of 30 and 100, respectively. These thresholds are intended to identify prominent local boundary transitions suitable for polygon refinement rather than to construct the flood-fill barriers used by Edge Fill. The optional snapping operation can improve the alignment of polygon vertices with locally detected image boundaries; however, its effectiveness depends on the correspondence between detected edges and the actual object boundary. In road imagery, nearby texture transitions, pavement markings, and other unrelated edges may occur close to the predicted contour. Snapping to such responses can introduce irregular or locally distorted polygon boundaries. For this reason, edge snapping is disabled by default and provided as an experimental refinement option for annotation cases in which closer pixel-level alignment with visible boundaries is desirable.

### 3.4 Pretrained Semantic Segmentation Suggestions

RoadAnnoter integrates six pretrained semantic segmentation models organized into two groups according to their training datasets, output class taxonomies, and inference architectures. The first group comprises three SegFormer variants fine-tuned on the Cityscapes benchmark, whereas the second includes three segmentation architectures available through the torchvision model library with COCO-trained weights and Pascal Visual Object Classes (VOC) output categories. The selection of a model group depends primarily on the annotation objective and the semantic categories represented in its output taxonomy. The Cityscapes-trained SegFormer models are recommended for road-scene annotation because their class taxonomy explicitly includes road surfaces, sidewalks, and several roadside infrastructure and environmental features. In contrast, the torchvision models provide categories suitable for identifying traffic participants and selected vehicles but do not include dedicated classes for road surfaces or roadside infrastructure. Consequently, their applicability is primarily limited to vehicle- and pedestrian-oriented annotation tasks rather than comprehensive road-environment characterization.

**SegFormer Cityscapes models (recommended for road-scene annotation).** Three SegFormer variants fine-tuned on the Cityscapes 19-class benchmark are integrated into RoadAnnoter: B0 (approximately 14 MB; 3.7 million parameters), B2 (approximately 25 MB; 27.4 million parameters), and B5 (approximately 84 MB; 84.6 million parameters). The models are loaded through the Hugging Face Transformers library (Xie et al., 2021; Cordts et al., 2016). The Cityscapes taxonomy comprises 19 semantic categories: road, sidewalk, building, wall, fence, pole, traffic light, traffic sign, vegetation, terrain, sky, person, rider, car, truck, bus, train, motorcycle, and bicycle. These categories provide broad coverage of the road environment, including carriageway surfaces, roadside surroundings, traffic-control infrastructure, and road users. This makes the Cityscapes-trained models particularly relevant to the annotation of road and roadside features in traffic safety research.

The three variants provide different trade-offs between computational requirements and segmentation performance. B0 has the smallest model size and is suitable for rapid inference and preliminary annotation previews. B2 provides an intermediate configuration that balances computational efficiency and segmentation capability for routine annotation tasks. B5 has the largest parameter count and generally achieves the highest segmentation performance among these three variants on the Cityscapes benchmark, making it suitable when segmentation quality is prioritized over computational efficiency. The model weights are obtained from the Hugging Face model hub and cached locally during the first use. Once the required weights have been downloaded, subsequent inference sessions can operate without an internet connection.

During inference, each input image is processed using the corresponding `SegformerImageProcessor`, which applies the required resizing, normalization, and input-format transformations. The processed image is then passed through the hierarchical Mix Transformer (MiT) encoder, which extracts multi-scale feature representations at spatial resolutions corresponding to 1/4, 1/8, 1/16, and 1/32 of the input dimensions. These feature representations are subsequently processed by the lightweight all-multilayer-perceptron (all-MLP) decoder, which projects and fuses the multi-scale information into a common feature representation. The resulting feature map is used to generate class-specific segmentation logits. These logits are bilinearly upsampled to the original image dimensions, and a channel-wise softmax operation converts them into per-pixel class probability distributions. The resulting probability maps provide the semantic predictions used by the subsequent suggestion-extraction procedure.

**torchvision COCO models.** The second group comprises three pretrained segmentation architectures: FCN with a ResNet-101 backbone, DeepLabV3 with a ResNet-101 backbone, and DeepLabV3 with a MobileNetV3-Large backbone. These models are loaded through the torchvision model library using the *COCO_WITH_VOC_LABELS_V1* pretrained weight configuration (Lin et al., 2014; Everingham et al., 2010). Their output taxonomy consists of 21 categories, comprising the 20 Pascal VOC object classes and a background category. The foreground categories include animals, aircraft, watercraft, persons, and several vehicle types, including bicycles, cars, motorbikes, buses, and trains. However, the taxonomy does not contain dedicated categories for road surfaces, sidewalks, terrain, or roadside infrastructure. Accordingly, these models cannot provide direct semantic segmentation of the principal road-environment classes required for comprehensive road and roadside feature annotation. Their inclusion nevertheless provides an additional annotation option for tasks focused on traffic participants and the vehicle categories represented in the VOC taxonomy.

Before inference, each image is resized to a dimension compatible with the model's spatial requirements. Specifically, the resized dimensions are selected as the nearest multiples of 32 within the range of 512–1024 pixels, using bilinear interpolation. The image is then normalized using the ImageNet channel statistics, with mean values of [0.485, 0.456, 0.406] and standard deviations of [0.229, 0.224, 0.225] for the red, green, and blue channels, respectively. The normalized image is passed through the corresponding segmentation architecture to generate class-specific logit maps. These maps are subsequently bilinearly upsampled to the original image dimensions, and a channel-wise softmax operation produces the per-pixel class probability distributions.

An important implementation distinction concerns the naming of these models within the application interface. Although the corresponding interface panel is labelled 'DeepLabV3+', the actual integrated architectures are FCN and DeepLabV3, rather than DeepLabV3+ as introduced by Chen et al. (2018). The interface label therefore does not accurately represent the complete set of underlying architectures. For methodological transparency and reproducibility, the manuscript identifies the models according to their actual torchvision implementations rather than the interface label.

**Suggestion extraction (common procedure for both model groups).** The per-pixel class probability maps generated by either model group are processed through a common suggestion-extraction procedure. First, the maximum class probability at each pixel is evaluated against a user-configurable confidence threshold, with a default value of 0.40. Pixels whose predicted class confidence falls below this threshold are suppressed and assigned to the background category. The remaining predictions are separated according to their respective predicted semantic classes and connected-component analysis is performed on each resulting binary class mask using eight-connectivity. This operation identifies spatially connected foreground regions and assigns a unique integer label to each component. The eight-connectivity criterion considers both horizontal and vertical neighbours, as well as diagonal neighbours, when determining whether foreground pixels belong to the same component.

Each connected component is subsequently evaluated against a minimum-area threshold of 2,000 pixels. Components smaller than this threshold are discarded to reduce the generation of insignificant regions and isolated segmentation artefacts. The remaining components are converted into polygon suggestions using the common mask-to-polygon conversion pipeline described in Section 3.6. The resulting polygons are inserted into the annotation suggestion queue and sorted in descending order of segmented area. This ordering prioritizes larger and more spatially prominent regions during the annotator's review, allowing substantial road-scene features to be considered before smaller components.

An important design consideration is the separation of model-predicted semantic categories from the project's annotation taxonomy. Although the pretrained models generate class-specific segmentation predictions, their output categories may not correspond directly to the classes defined in the annotation project. Furthermore, differences between the models' training datasets and the target road-scene imagery may introduce domain-shift effects that reduce the reliability of predicted class assignments. To avoid automatically transferring potentially inconsistent semantic labels into the project annotations, RoadAnnoter assigns class index 0 to every generated suggestion as an initial placeholder, irrespective of the model-predicted category. The annotator is therefore required to inspect each suggestion and explicitly assign the appropriate project-specific class before accepting it. This design preserves the utility of pretrained models for identifying candidate object boundaries while retaining human control over the final semantic classification.

## 3.5 Lightweight Built-In Segmentation Model (TinySegNet)

**Architecture.** TinySegNet is a lightweight semantic segmentation network developed to generate annotation suggestions using the project's own labelled images. Its encoder is based on MobileNetV3-Small (Howard et al., 2019), initialized with ImageNet-pretrained weights to provide transferable visual representations and reduce the amount of task-specific training required. MobileNetV3-Small incorporates inverted residual blocks with linear bottlenecks, hard-swish activation functions, and squeeze-and-excitation modules for channel-wise feature recalibration. These architectural components provide an efficient balance between representational capacity and computational complexity, making the encoder suitable for incremental model development on relatively small annotated datasets and resource-constrained computing environments.

The encoder extracts a 576-channel feature representation at 1/16 of the input spatial resolution. Thus, for the default input size of 512 × 512 pixels, the resulting feature map has spatial dimensions of 32 × 32 pixels. This representation provides a compact encoding of the input image while retaining spatial information for subsequent segmentation. However, the degree of boundary detail preserved in the final prediction also depends on the decoder architecture and the subsequent upsampling operation.

To incorporate contextual information at multiple spatial scales, the encoder output is passed through a lightweight Atrous Spatial Pyramid Pooling (ASPP) module based on the multi-scale context aggregation strategy described by Chen et al. (2018). The module consists of four parallel processing branches. The first branch applies a 1 × 1 convolution to capture local channel relationships without expanding the spatial receptive field. The second and third branches employ 3 × 3 atrous convolutions with dilation rates of r = 6 and r = 12, respectively. These operations expand the effective receptive fields to 13 × 13 and 25 × 25 feature-map positions, respectively. Given the assumed encoder output stride of 16, these correspond to approximate input-image receptive-field extents of 208 × 208 and 400 × 400 pixels. The fourth branch performs global average pooling, followed by a 1 × 1 convolution and bilinear upsampling to the spatial dimensions of the encoder feature map. This branch introduces image-level contextual information that complements the spatially localized features extracted by the other branches.

Each ASPP branch produces a 128-channel feature representation. The four outputs are concatenated to form a 512-channel feature tensor, which is subsequently projected to 128 channels using a 1 × 1 convolution followed by batch normalization, ReLU activation, and dropout with a probability of 0.1. A subsequent 3 × 3 convolution refines the fused representation to 64 channels. Finally, a 1 × 1 classification layer generates pixel-wise segmentation logits over N + 1 output channels, where N represents the number of project-specific foreground classes and the additional channel represents the background category. The resulting logits are bilinearly upsampled by a factor of 16 to restore the predictions to the original input resolution. This encoder–context aggregation–classification architecture provides a compact segmentation network that can be trained using annotations accumulated during the normal operation of RoadAnnoter.

**Training.** TinySegNet is trained using the polygon annotations available in the current project folder. The training dataset is assembled dynamically from images associated with non-empty annotation files, allowing newly accepted annotations to become available for subsequent training sessions. The current implementation does not create separate training, validation, or test subsets; therefore, all available labelled images contribute to model optimization. This design facilitates rapid adaptation during annotation development but does not provide an independent dataset for evaluating generalization performance.

Ground-truth segmentation masks are generated by rasterizing the polygon annotations using OpenCV's `fillPoly` function with 32-bit integer vertex coordinates. Each polygon is filled with its corresponding model-training class label, while pixels outside annotated polygons are assigned to the background category. When polygons overlap, the annotation appearing later in the label-file sequence takes precedence because rasterization is performed sequentially in file order. To maintain consistency between image and mask dimensions, input images are resized to a configurable square resolution, with 512 × 512 pixels as the default, using bilinear interpolation. The corresponding segmentation masks are resized using nearest-neighbour interpolation to preserve discrete class labels and prevent the introduction of artificial intermediate class values at annotation boundaries. No data augmentation is applied in the current implementation.

The network is optimized using the multi-class cross-entropy loss and the AdamW optimizer with a weight-decay coefficient of $1 \times 10^{-4}$. AdamW decouples weight decay from the adaptive gradient-based parameter updates, providing explicit weight regularization during optimization. A cosine annealing learning-rate scheduler progressively reduces the learning rate from its initial value towards the end-of-training value over the specified training period. The default training configuration consists of a batch size of 4 and 20 epochs, although both parameters can be adjusted by the user.

Training is executed in a background `QThread`, allowing the graphical annotation interface and image canvas to remain responsive while model optimization proceeds. The training process emits per-epoch progress information, including the epoch index, mean batch loss, and training mean intersection over union (mIoU). The mIoU is calculated over the foreground classes, excluding the background category. Two model checkpoints are maintained during training: *`best_seg.pt`*, which stores the checkpoint associated with the highest observed training mIoU, and *`last_seg.pt`*, which stores the most recently completed epoch's checkpoint. Because checkpoint selection and reported mIoU are based exclusively on training data, these values indicate optimization progress and convergence behaviour rather than independent segmentation accuracy or generalization performance.

**Own-model suggestions.** Once a trained TinySegNet checkpoint is loaded, the model generates a dense semantic prediction for an input image through a single forward inference pass. The resulting class map is processed separately for each foreground class, excluding the background category. For each class, a binary foreground mask is constructed and subjected to connected-component analysis to identify spatially distinct predicted regions. Each connected component is evaluated according to its polygon area, and components exceeding the minimum area threshold of 100 pixels are converted into polygon suggestions through the shared mask-to-polygon conversion pipeline described in Section 3.6. Unlike suggestions generated by the pretrained semantic segmentation models, TinySegNet suggestions retain their predicted project-specific class indices. This is possible because the model is trained using the annotation taxonomy defined within the current project, allowing its predicted categories to be interpreted directly in the project's semantic classification framework.

The generated polygons are added to the annotation suggestion queue, where the annotator reviews their spatial boundaries and predicted class assignments before acceptance. Correct suggestions can be accepted directly, whereas inaccurate predictions can be rejected or corrected through the annotation interface. Accepted and corrected annotations subsequently become part of the available project-labelled dataset and can be incorporated into later training sessions. Through repeated annotation, dataset expansion, and model retraining, TinySegNet supports an iterative human-in-the-loop learning process in which the accumulated project-specific annotations progressively provide additional supervision for improving subsequent segmentation suggestions. The effectiveness of this process depends on the quality, diversity, and representativeness of the accumulated annotations, as well as the training configuration and frequency of model updates.

### 3.6 Shared Mask-to-Polygon Conversion Pipeline

All five assisted annotation modes generate binary segmentation masks that are processed through a common six-stage mask-to-polygon conversion pipeline before being presented to the annotator as polygon suggestions. These modes include Edge Fill, SAM Prompt, Smart Selection, pretrained semantic segmentation suggestions, and TinySegNet-based suggestions. The conversion procedure transforms binary masks, which may contain irregular boundaries, internal holes, isolated foreground pixels, or multiple disconnected regions, into simplified polygon representations suitable for visual inspection and storage in the YOLO segmentation format. Applying a common conversion procedure across all five modes establishes a consistent geometric representation of their segmentation outputs and ensures that modifications to the conversion pipeline can be applied uniformly, irrespective of the segmentation method that generated the original mask.

**Step 1: Morphological smoothing.** The initial binary mask is subjected to morphological closing using an $11 \times 11$ elliptical structuring element, followed by morphological opening using a $5 \times 5$ elliptical structuring element. Closing, which consists of dilation followed by erosion, bridges narrow discontinuities and fills sufficiently small internal gaps within the foreground region. Such discontinuities may arise from background pixels enclosed within the segmented region, local variations in colour similarity, or pixels suppressed by the confidence threshold during deep-learning-based segmentation. Opening, which consists of erosion followed by dilation, subsequently removes small isolated foreground components and narrow protrusions along the region boundary. The combined operations reduce local irregularities and produce a more spatially coherent mask before contour extraction. The actual extent of gap filling and protrusion removal depends on the structuring-element geometry and the local shape of the segmented region.

**Step 2: Gaussian smoothing and re-thresholding.** The morphologically refined binary mask is convolved with a $21 \times 21$ Gaussian kernel to smooth local boundary irregularities and reduce the staircase effects associated with pixel-grid-based segmentation boundaries. Gaussian filtering produces intermediate intensity values around foreground–background transitions, effectively creating a gradual transition between the two regions. The filtered mask is subsequently converted back into a binary representation by applying a fixed threshold of 127. Pixels with filtered intensities above the threshold are retained as foreground, whereas the remaining pixels are assigned to the background. This operation produces a smoothed binary boundary that reduces small pixel-level fluctuations and provides a more regular contour for subsequent polygon extraction. The degree of boundary smoothing is controlled by the Gaussian kernel size and the re-thresholding operation.

**Step 3: Contour extraction.** The smoothed binary mask is processed using OpenCV's `cv2.findContours` function with the *`RETR_EXTERNAL`* retrieval mode. This mode extracts the external boundary of each connected foreground component while excluding interior contours associated with enclosed background regions. The contour extraction procedure follows the border-following approach described by Suzuki and Abe (1985). When multiple external contours are returned, the contour with the largest enclosed area is retained, whereas all smaller contours are discarded. Consequently, the resulting representation consists of a single external contour corresponding to the largest detected foreground region. This operation provides a consistent basis for polygon generation but also introduces a geometric limitation: smaller disconnected regions and internal holes are not preserved in the resulting polygon representation.

**Step 4: Optional edge snapping (SAM-based modes only; disabled by default).** For SAM Prompt and Smart Selection, an optional edge-snapping operation can be applied to the extracted contour before polygon simplification. When enabled, individual contour vertices are examined for nearby Canny edge responses within a configurable search radius. A vertex may be relocated to the nearest eligible edge pixel, subject to the constraint that the candidate location must remain within the original SAM mask boundary. If no eligible edge response is identified within the search neighbourhood, the vertex retains its original position. The operation uses an independently generated Canny edge map and is intended to improve the alignment of polygon vertices with locally detected image boundaries. However, because nearby image edges may correspond to internal textures or unrelated objects rather than the target boundary, snapping can introduce local contour irregularities. It is therefore disabled by default and is available as an optional refinement procedure, as described in Section 3.3. This stage is skipped for Edge Fill, pretrained semantic segmentation suggestions, and TinySegNet-based suggestions.

**Step 5: Douglas–Peucker polygon simplification.** The extracted contour is simplified using the Douglas–Peucker algorithm (Douglas & Peucker, 1973) to reduce the number of vertices while retaining the principal geometric characteristics of the boundary. First, the contour perimeter is calculated, and the simplification tolerance is defined as 0.3% of the perimeter **$\varepsilon = 0.003 \times P$** where $\varepsilon$ denotes the simplification tolerance in pixels and P represents the contour perimeter in pixels. The Douglas–Peucker algorithm recursively approximates the contour using line segments. At each iteration, the perpendicular distances of intermediate contour points from the line segment connecting the endpoints are evaluated. Points whose maximum perpendicular distance exceeds ε are retained, and the procedure is recursively applied to the resulting subsegments. Points falling within the specified tolerance are removed. This process converts the original pixel-level contour, which may contain hundreds of boundary coordinates, into a substantially smaller set of polygon vertices. The resulting reduction in vertex count decreases annotation-file complexity and limits the retention of minor boundary fluctuations, while maintaining an approximation of the principal object geometry. The degree of geometric simplification is governed by the perimeter-relative tolerance and therefore varies with the size and complexity of the extracted contour.

**Step 6: Polygon validity check.** The simplified contour is subjected to a final geometric validity check based on its vertex count. Any result containing fewer than three distinct vertices is discarded because it cannot define a valid closed polygon. Contours satisfying this minimum requirement are retained and converted into polygon suggestions for presentation within the annotation interface. The resulting polygon is subsequently available for manual inspection and correction before acceptance and storage.

Two geometric limitations arise directly from the design of this conversion pipeline and should be considered when interpreting the resulting annotations. First, interior holes are not preserved in the polygon representation because the `RETR_EXTERNAL` contour retrieval mode retains only external boundaries. Consequently, enclosed background regions, such as the sky visible through a bridge arch or a pothole surrounded by a segmented road surface, are incorporated into the polygon's interior and represented as foreground. This can lead to an overestimation of the segmented area when the target object contains substantial internal background regions. Second, when a mask contains multiple disconnected foreground components, retaining only the largest external contour results in the exclusion of smaller components during polygon generation. This limitation is particularly relevant to masks containing spatially separated features or fragmented predictions. The pretrained semantic segmentation and TinySegNet suggestion modes partially address the second limitation by performing connected-component analysis before invoking the shared conversion pipeline. Each qualifying component is processed separately and can therefore generate an independent polygon suggestion, subject to the corresponding minimum-area threshold. However, the pipeline itself continues to retain only the largest external contour within each individual component mask.

## 3.7 Label Storage Format

Accepted polygon annotations are stored in the YOLO segmentation format, in which each annotation occupies a separate line in a text-based label file. Each line consists of an integer class identifier followed by alternating normalized horizontal and vertical polygon coordinates, represented in the following structure: $class_{id}\ x_1\ y_1\ x_2\ y_2 \ldots x_n\ y_n$.

Here, *class_id* is a zero-indexed integer corresponding to the position of the annotation class in the project-wide classes.txt file, while ($x_i$, $y_i$) represents the normalized coordinates of the $i$th polygon vertex. Pixel coordinates are normalized independently by the image width ( $x_{i norm} = x_{i pixel} / W$ ) and height ($y_{i norm} = y_{i pixel} / H$) .

where $W$ and $H$ denote the original image width and height, respectively. The normalized coordinates are constrained to the image coordinate range and written to six decimal places. This precision provides a sufficiently fine numerical representation of polygon vertices for typical road-scene image dimensions while maintaining a compact text-based annotation format. Each polygon is stored as a sequence of vertex-coordinate pairs following its corresponding class identifier.

The label file associated with each image is stored in the same project directory and shares the image's base filename, with a .txt extension. For example, an image named *image_001.jpg* has a corresponding annotation file named *image_001.txt*. When an annotation is accepted, its polygon coordinates and class identifier are written to the corresponding label file. Confidence scores, model predictions, and temporary suggestion-state information are not included in the stored annotation records; these attributes are maintained only during the active annotation session and are discarded upon acceptance.

To reduce the risk of partially written annotation files, RoadAnnoter employs an atomic file-replacement procedure. The updated annotation content is first written to a temporary file, after which os.replace is used to replace the existing label file with the completed version. This approach reduces the possibility of exposing an incomplete label file if the application is interrupted during the writing process, subject to the atomic-replacement guarantees of the underlying filesystem.

Class definitions are maintained separately in a project-wide classes.txt file, with one class name recorded per line. The position of each class name in this file determines its corresponding zero-indexed class identifier. Consequently, the integer class identifier stored in each annotation line is resolved to its semantic class name through the current class list when annotations are displayed in the application. This separation between numerical class identifiers and their textual definitions provides a centralized mechanism for managing project-specific annotation categories.

In addition to the image and label files, RoadAnnoter uses an optional zero-byte .reviewed marker file to record the review status of individual images. The marker is generated using the corresponding image filename with the .reviewed extension appended. For example, the review status of *image_001.jpg* is represented by *image_001.jpg.reviewed*. Because this marker is maintained independently of the polygon label file, image-review status can be recorded without modifying the underlying annotation content. This separation allows the application to manage annotation data and image-level review status independently, supporting the organization and tracking of annotation progress across a project.

# 4 Data Collection and Processing

## 4.1 Image Acquisition

Street-view images were acquired from recorded road accident locations across Bangladesh using the Google Street View Static API. For each accident record, the associated GPS coordinates were submitted to the API to retrieve a georeferenced, street-level photograph representing the road environment at the recorded location. The images were downloaded programmatically and stored locally for subsequent annotation and processing.

Google Street View imagery was selected because it enables the systematic acquisition of georeferenced road-scene photographs across a large number of accident locations without requiring extensive field surveys. This approach offers practical advantages in terms of data collection time, logistical requirements, and cost, particularly when investigating geographically dispersed accident locations across an extensive road network. Furthermore, the predominantly forward-facing street-level perspective provides a broadly comparable visual configuration to that of established road-scene datasets, such as Cityscapes (Cordts et al., 2016). This similarity in viewing perspective facilitates the application of segmentation models pretrained on Cityscapes, although differences in geographical context, road infrastructure, traffic composition, and image characteristics may still introduce domain-shift effects.

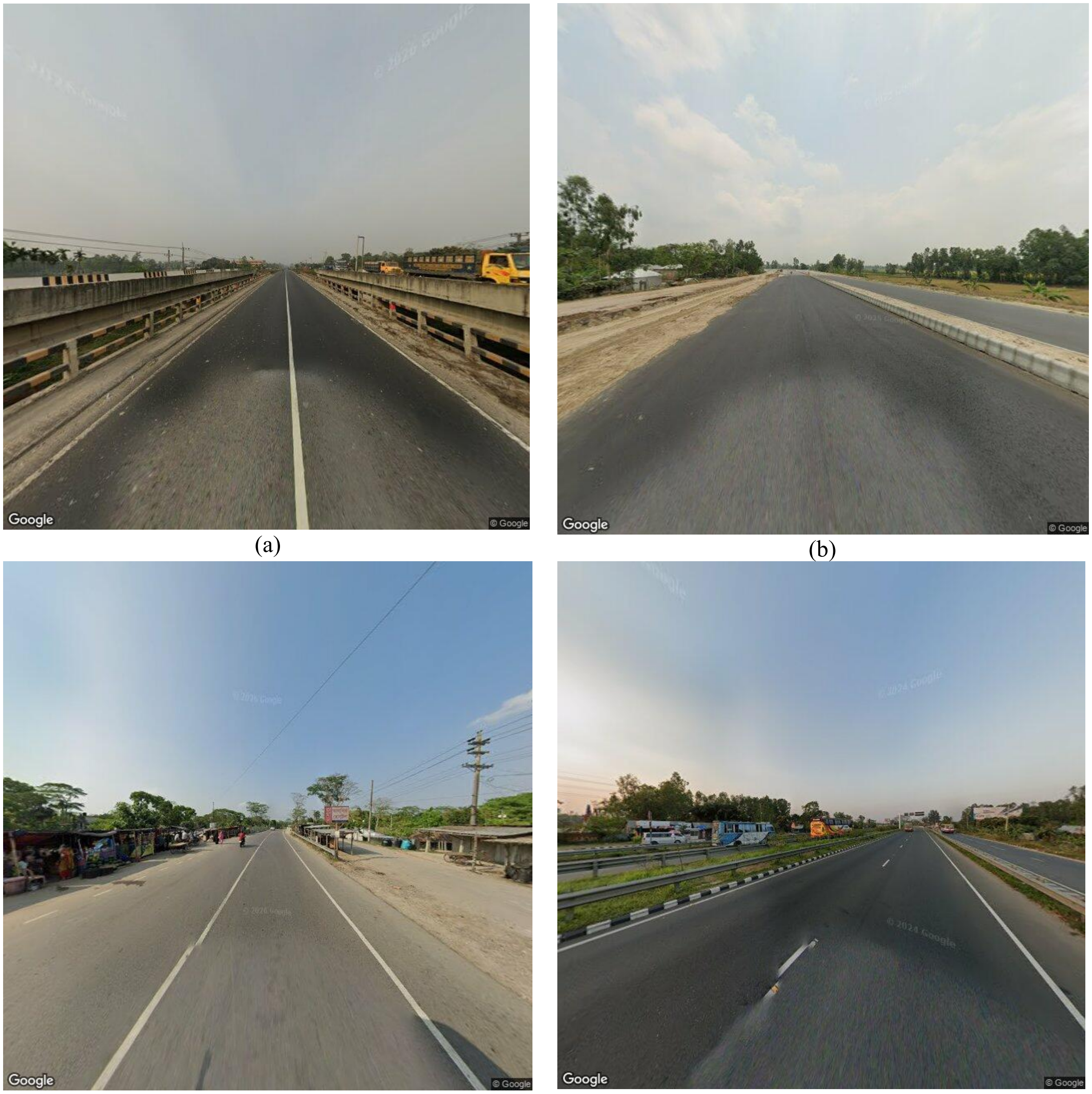

(a) (b) (c) (d)

**Figure 2** Photos Collected using Google Static Street View Imagery

## 5 Visualization of Annotation Procedure

This section presents representative annotation outputs generated using the RoadAnnoter interface across four complementary annotation modes: manual polygon drawing, SAM Prompt, Smart Selection, and pretrained model suggestions. The examples are drawn from road-scene images acquired through Google Street View at recorded road accident locations in Bangladesh. Collectively, they demonstrate the applicability of the annotation framework to diverse feature geometries encountered in road safety research, ranging from compact infrastructure elements to horizontally elongated road features and large, irregular surface regions.

Figure 3 illustrates the manual polygon annotation mode, in which the annotator delineates an object by placing successive vertices on the image canvas and closing the polygon by pressing Enter or double-clicking the initial vertex. This mode does not require a deep learning model and can therefore be applied to objects of arbitrary shape, provided that their boundaries can be visually identified. The resulting annotation quality depends primarily on the accuracy of the annotator's vertex placement. Figure 3(a) presents an example of a relatively regular, approximately convex feature, such as a road sign or a small infrastructure element, for which a limited number of vertices can adequately represent the boundary. In contrast, Figure 3(b) illustrates an irregular, non-convex feature, such as a pavement patch or roadside vegetation strip, which requires additional vertices to capture its boundary geometry. During export, the polygon is simplified using the Douglas–Peucker algorithm, with a tolerance set to 0.3% of the polygon perimeter, to reduce the number of vertices while retaining the principal boundary characteristics.

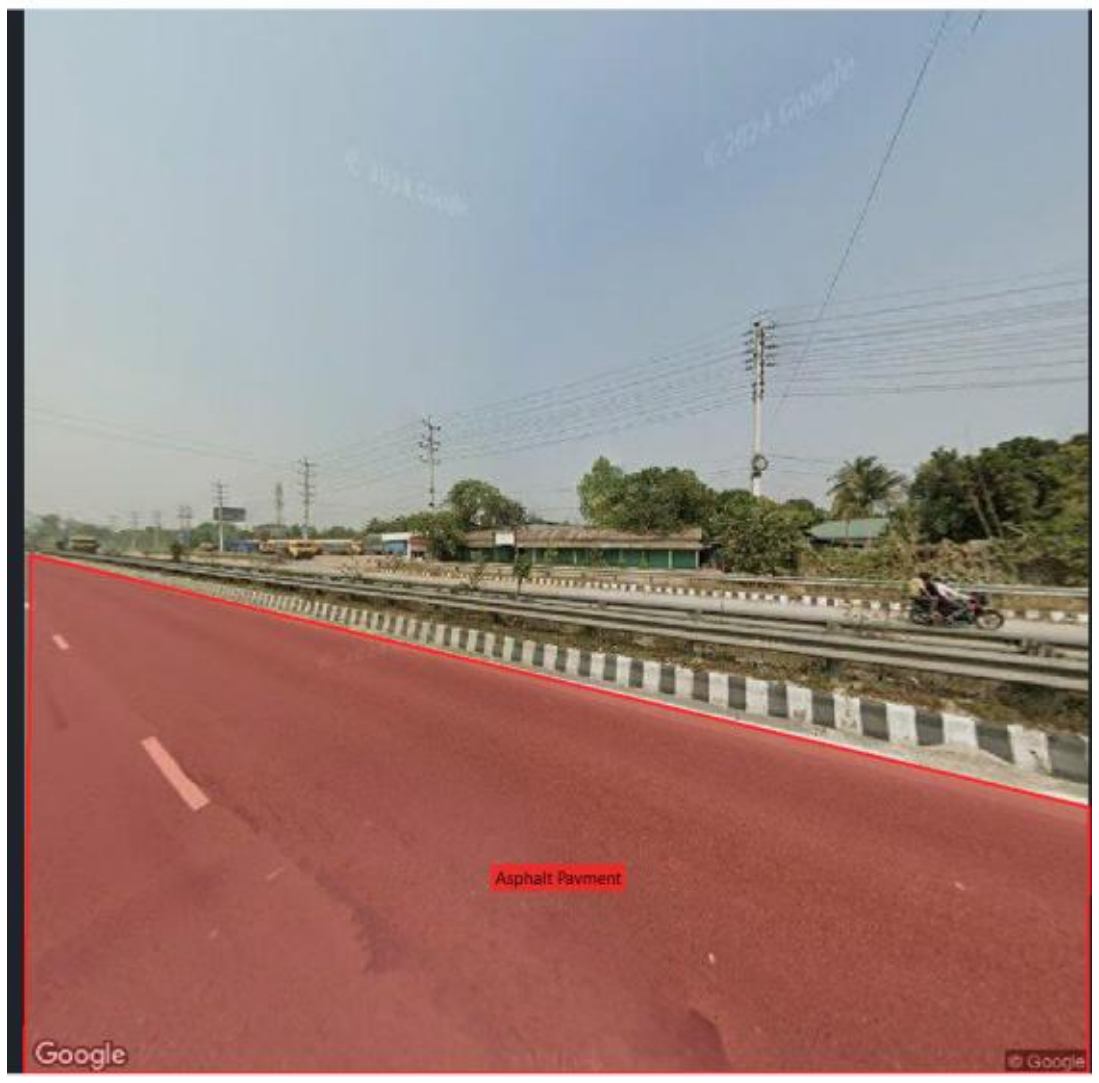

(a) Regular Shape Selection

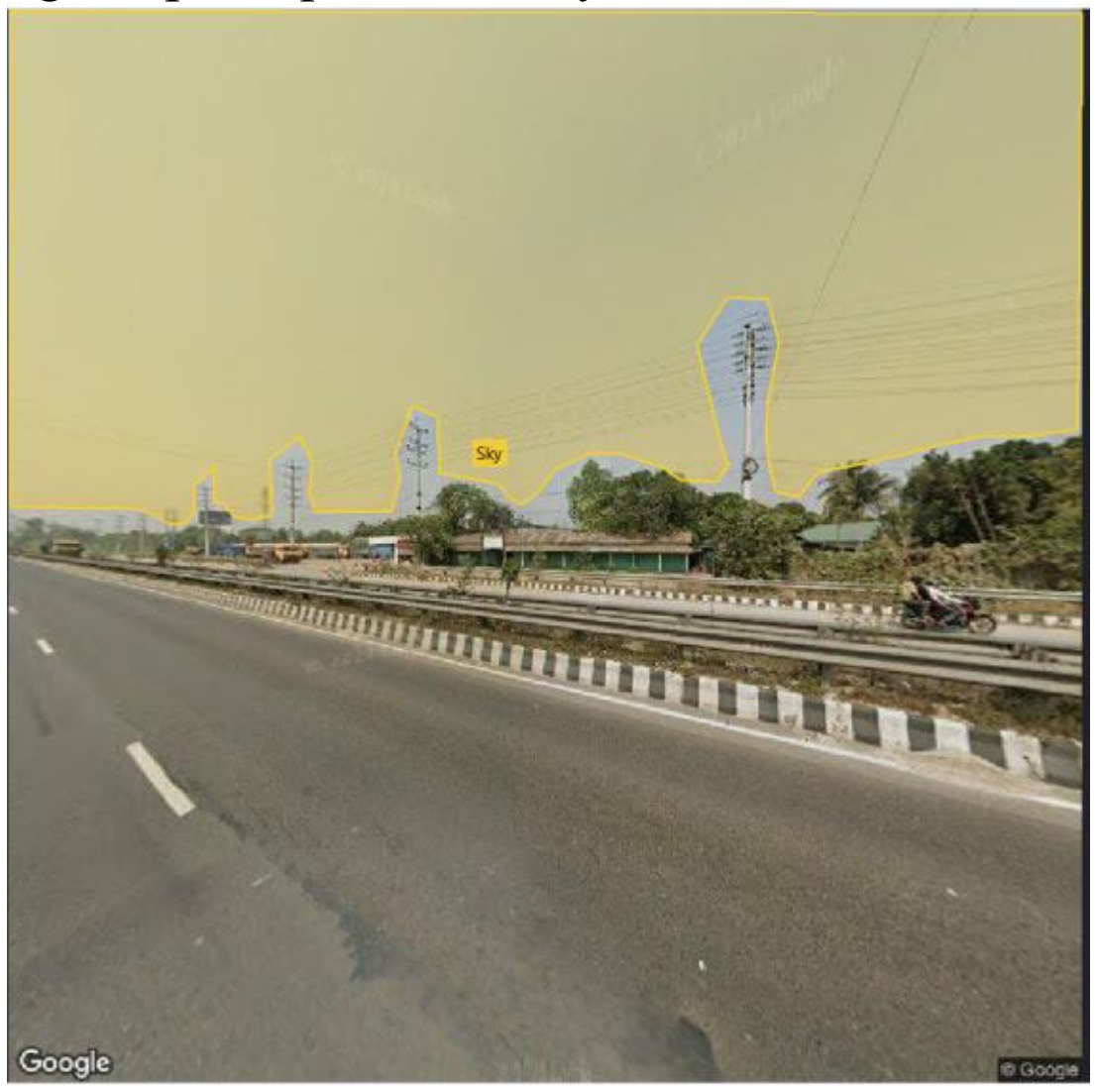

(b) Irregular Shape Slection

**Figure 3** Different Features of Polygon Selection and Annotation (mouse action- drawing)

**Figure 4** demonstrates the SAM Prompt annotation mode. In this mode, the annotator places one or more foreground point prompts on the target object using the left mouse button. Background exclusion points can also be introduced using the right mouse button to discourage the segmentation model from including irrelevant regions. The Segment Anything Model (SAM) (Kirillov et al., 2023) processes the accumulated prompts together with the precomputed 256-channel image embedding to generate three candidate binary masks. The candidate with the highest predicted intersection-over-union (IoU) score is selected and subsequently processed through the shared mask-to-polygon conversion pipeline. The annotator can accept the proposed polygon or refine the segmentation by introducing additional prompts. The four panels illustrate the mode's applicability to representative geometric categories encountered in road-scene annotation: (a) irregular, non-convex surfaces, such as road damage patches; (b) compact features with relatively regular boundaries, such as road signs or drain covers; (c) horizontally elongated features, including curb lines, lane markings, and guardrail sections; and (d) vertically oriented structures, such as utility poles and tree trunks. These examples demonstrate how point-based prompting can facilitate polygon generation across different object geometries, potentially reducing the need for manually placing individual polygon vertices.

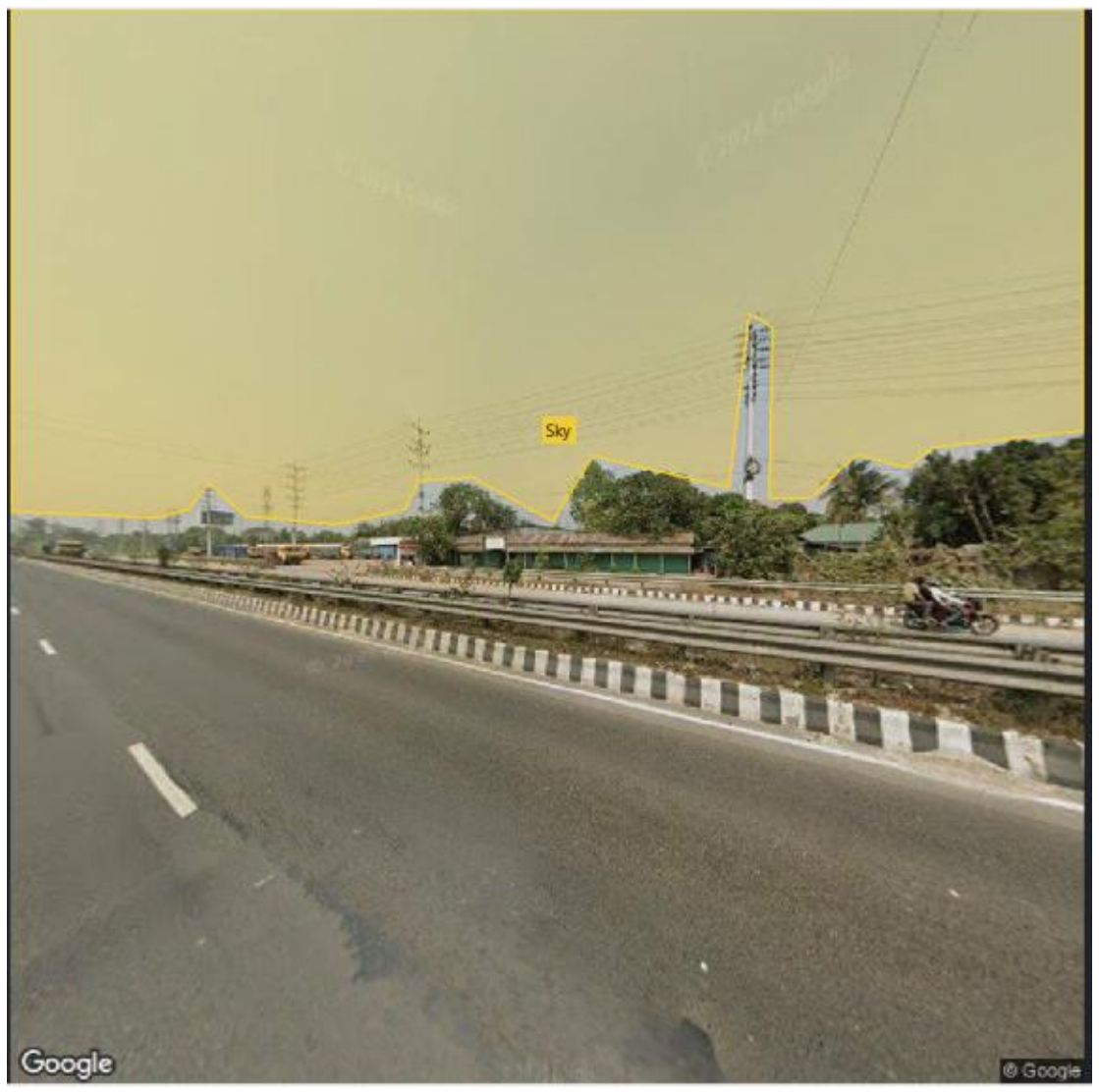

(a) Irregular  Shape

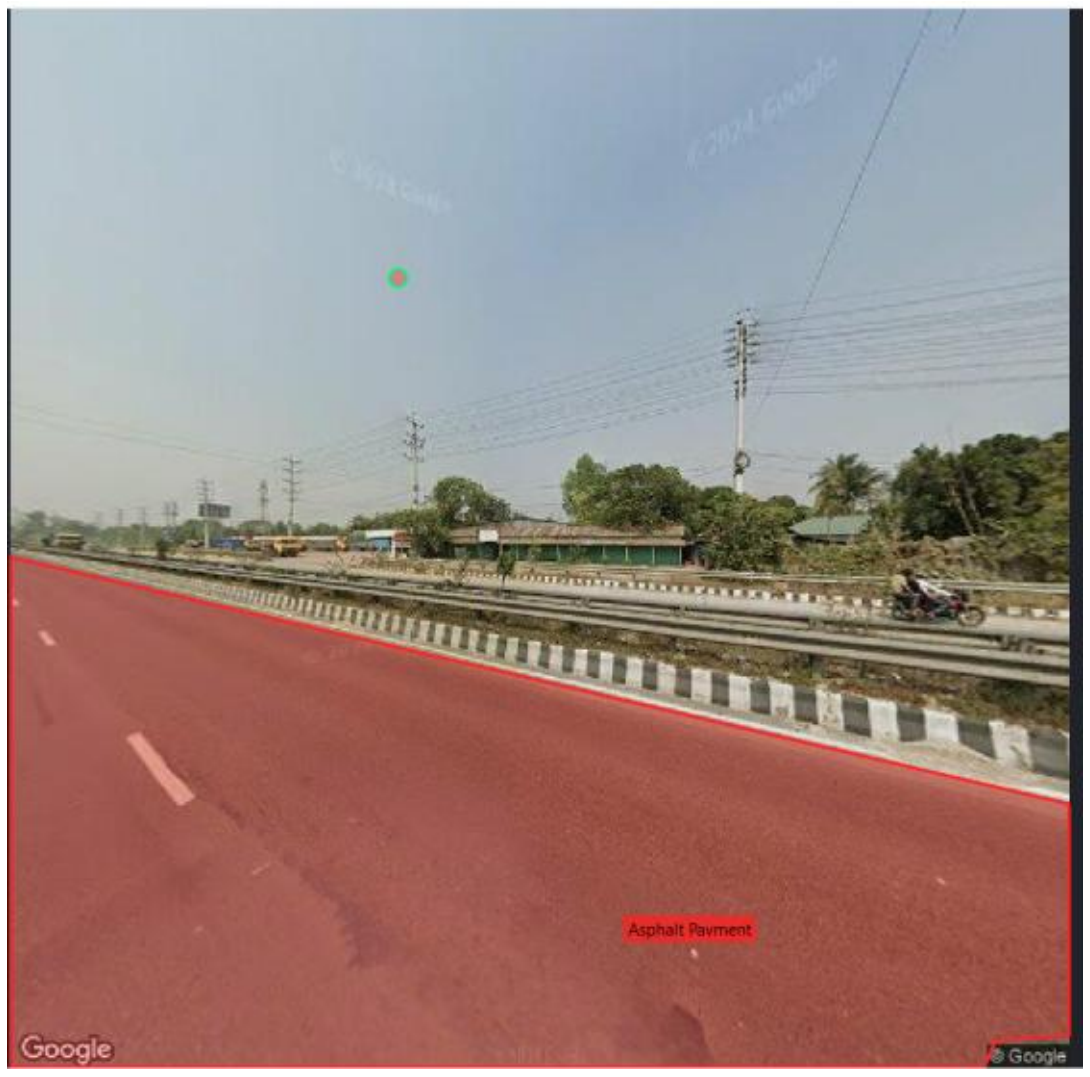

(b) Regular Shape

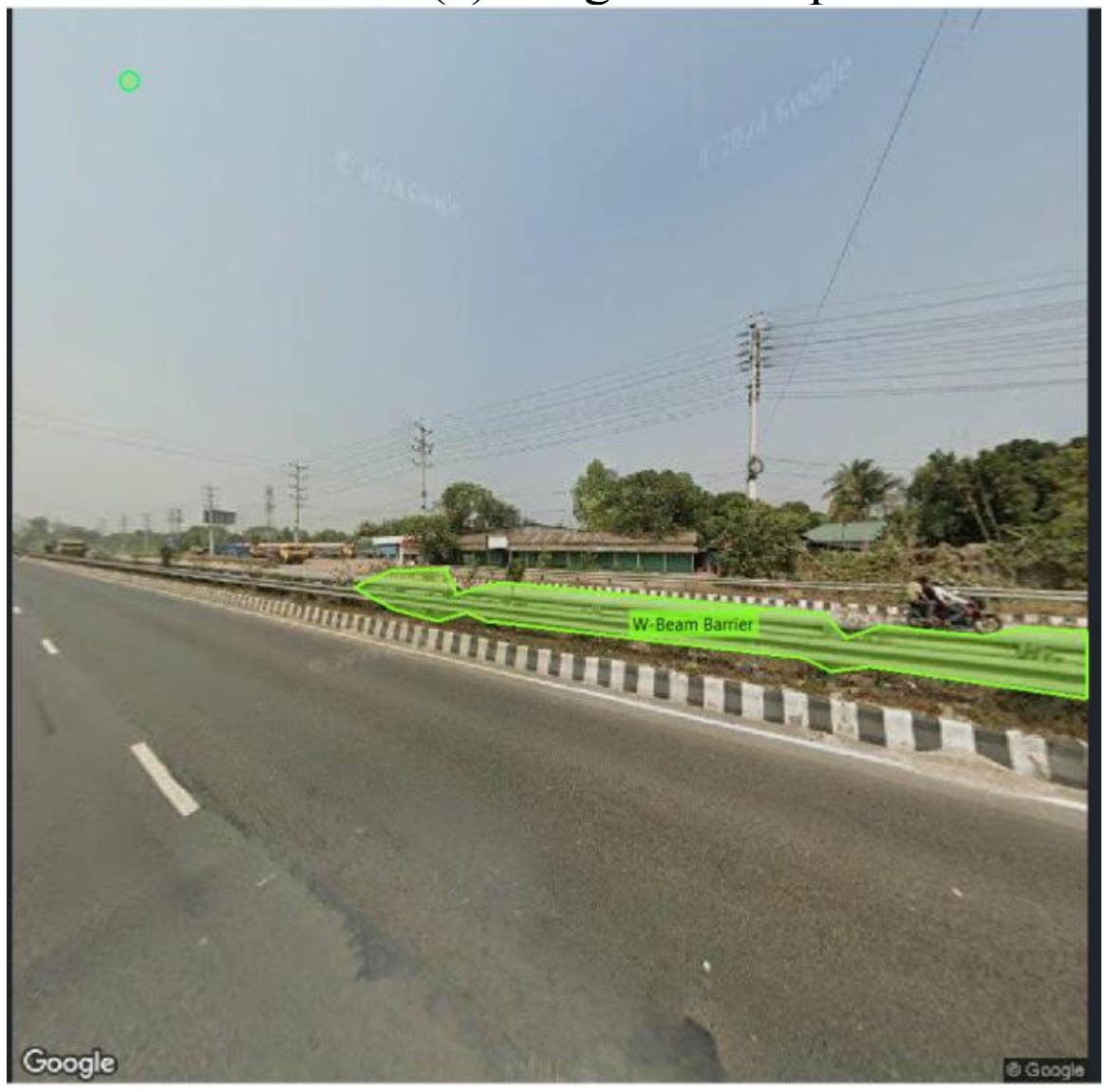

(c) Horizontally Long Shape

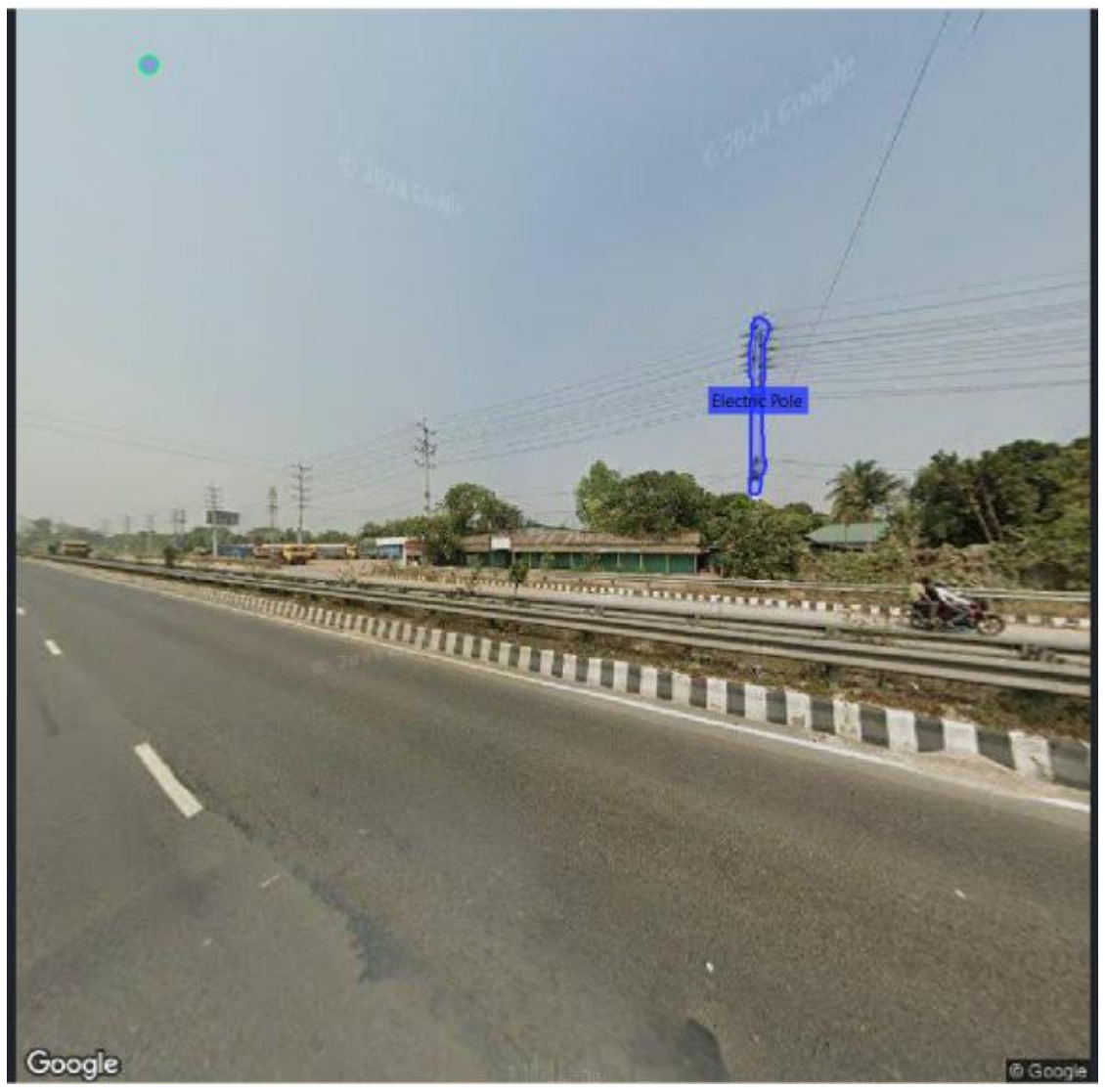

(d) Vertically Long Shape

**Figure 4** Different Features of SAM based Selection and Annotation (mouse action-clicking)

**Figure 5** illustrates the Smart Selection mode, which implements a one-click variant of the SAM-based annotation procedure. A single left-click on the target feature generates a combined point-and-box prompt. The bounding box is centred on the selected point and extends by the configured 40% of the image dimensions in each direction, subject to the image boundaries. SAM generates three candidate masks from this combined prompt, after which the candidate with the largest area is selected. This selection criterion prioritises region coverage rather than the predicted IoU score and is intended to facilitate the annotation of road surfaces and relatively large roadside features for which incomplete region coverage may be undesirable. The annotator can review and accept the resulting polygon without introducing additional prompts when the initial segmentation is satisfactory. The four panels show Smart Selection outputs for the same broad geometric categories presented in Figure 4: (a) regular-shaped objects, (b) irregular and non-convex surfaces, (c) horizontally elongated features, and (d) vertically oriented structures. A comparison of Figures 4 and 5 highlights the distinction between the two interaction strategies. SAM Prompt provides greater control through iterative foreground and background prompting, whereas Smart Selection prioritises interaction speed by generating a segmentation proposal from a single click. The two modes therefore offer complementary approaches to prompt-based annotation, with their relative suitability depending on the geometry and segmentation requirements of the target feature.

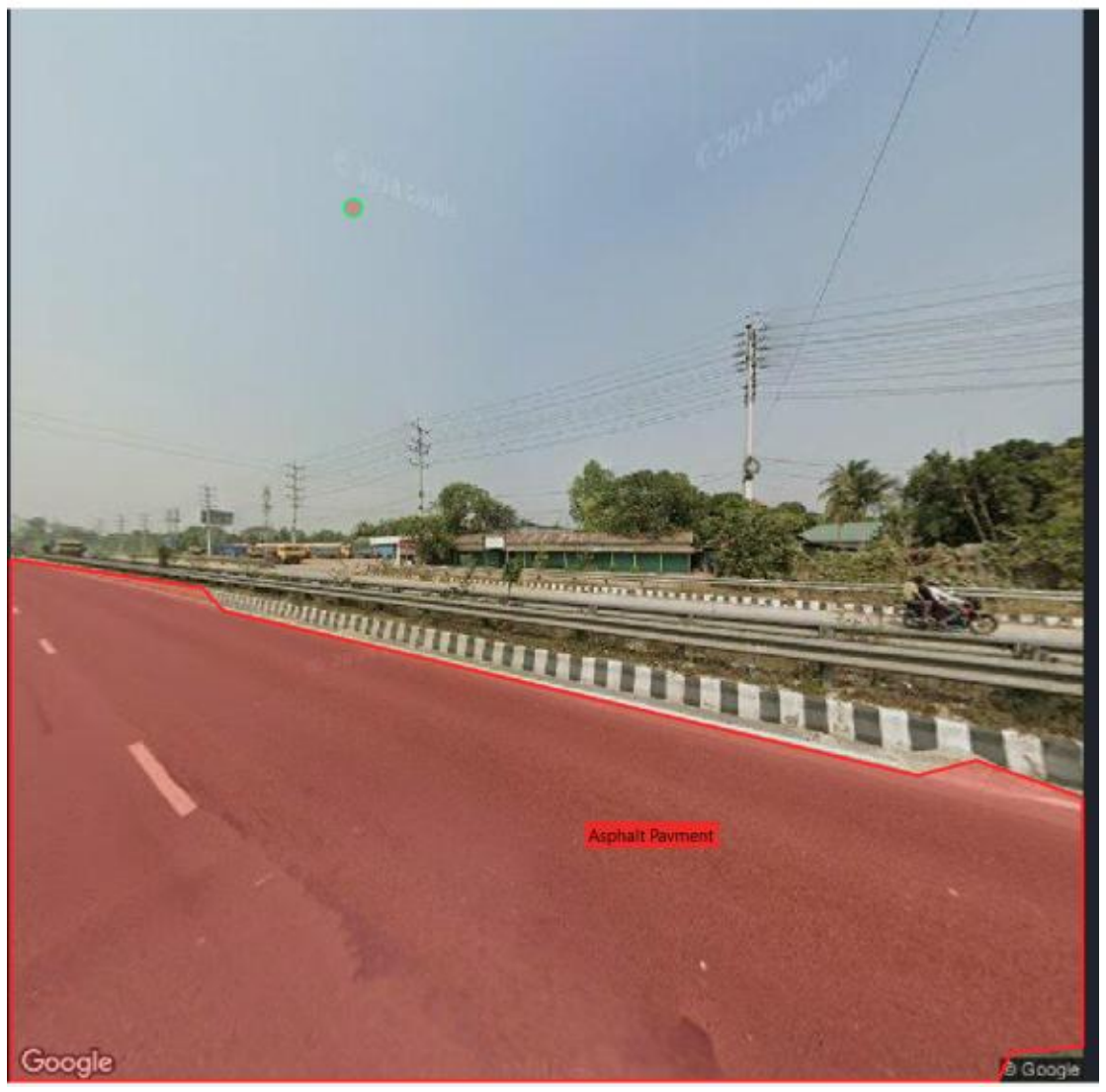

(a) Regular Shape

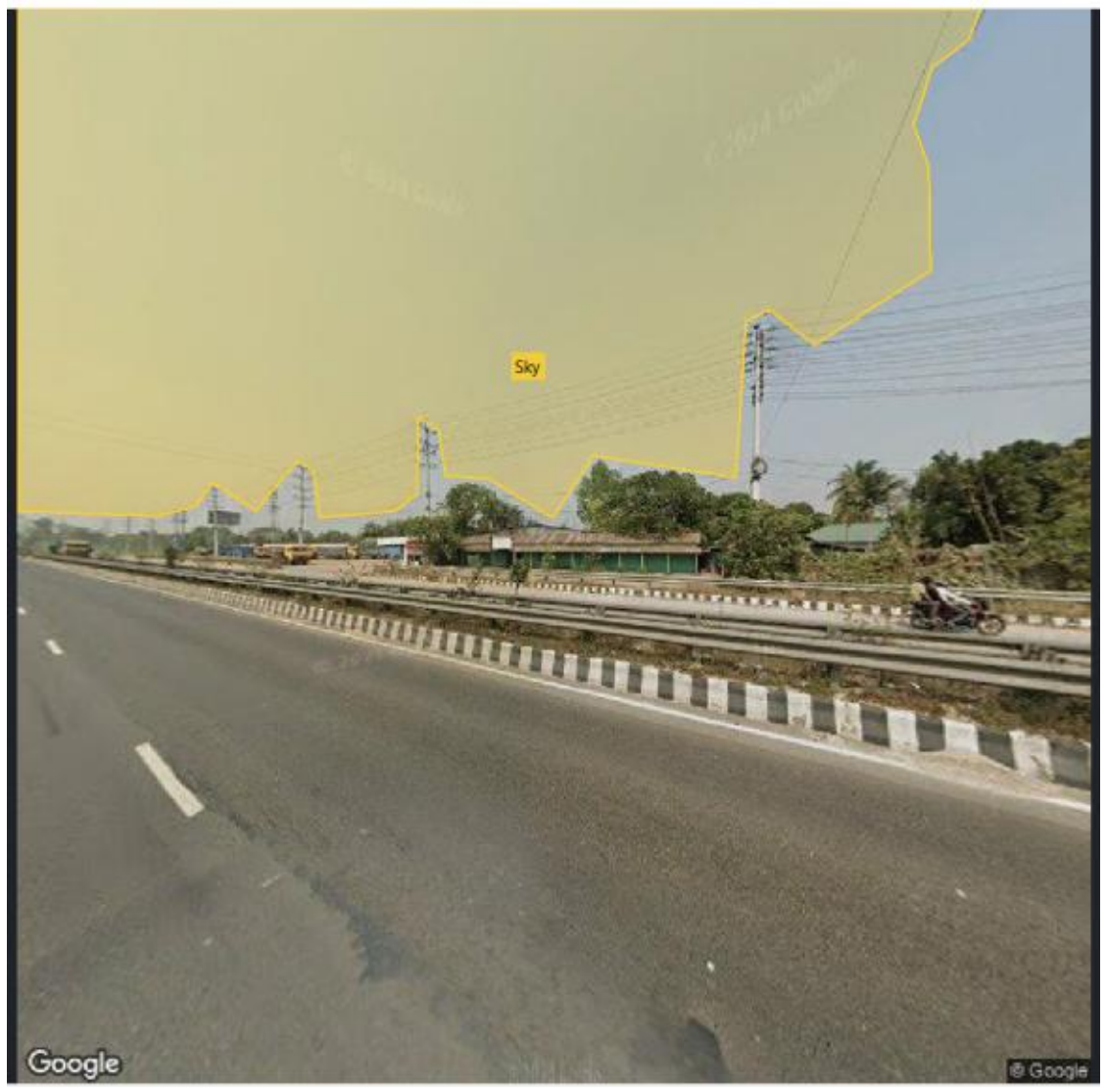

(b) Irregular  Shape

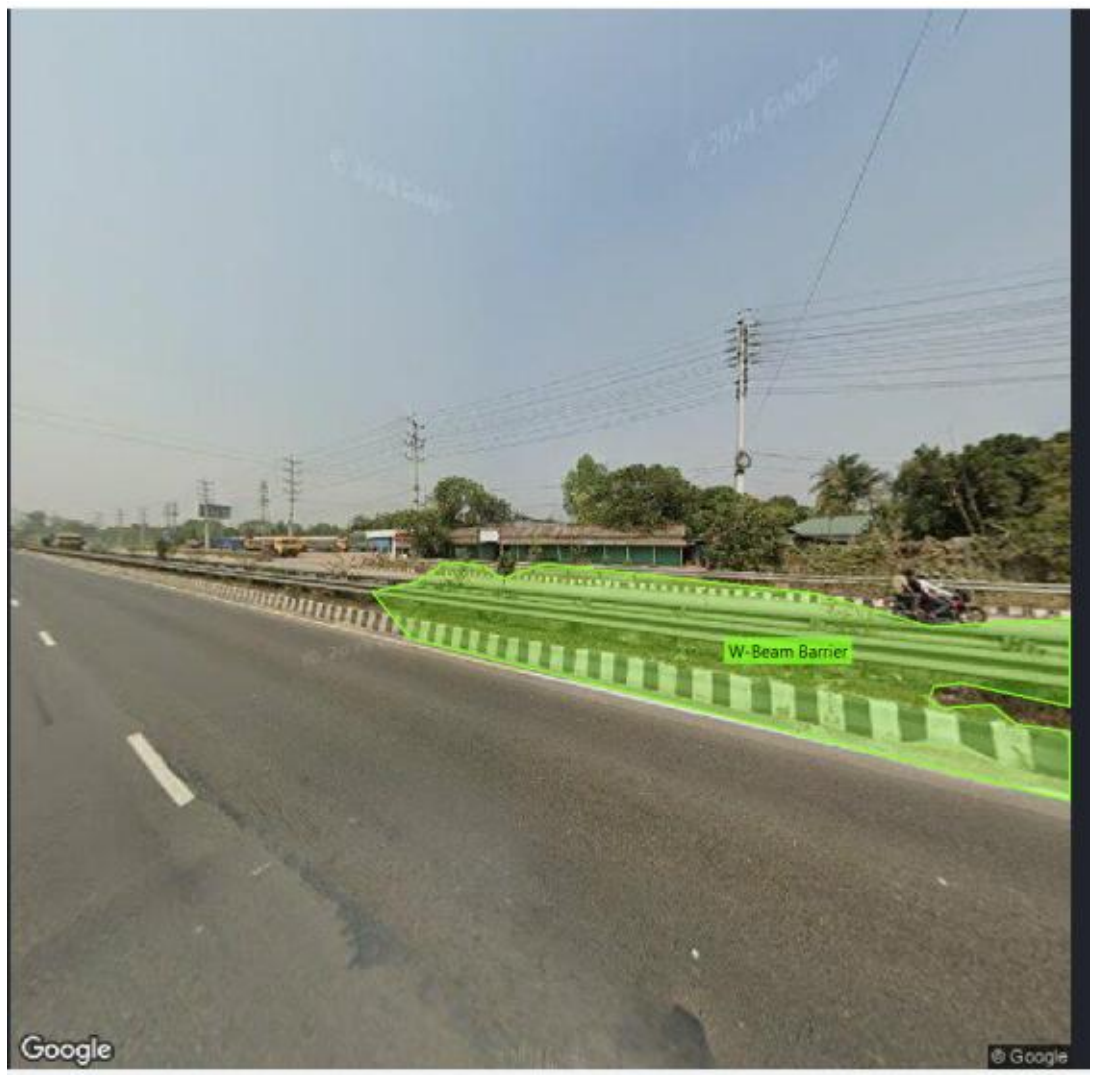

(c) Horizontally Long Shape

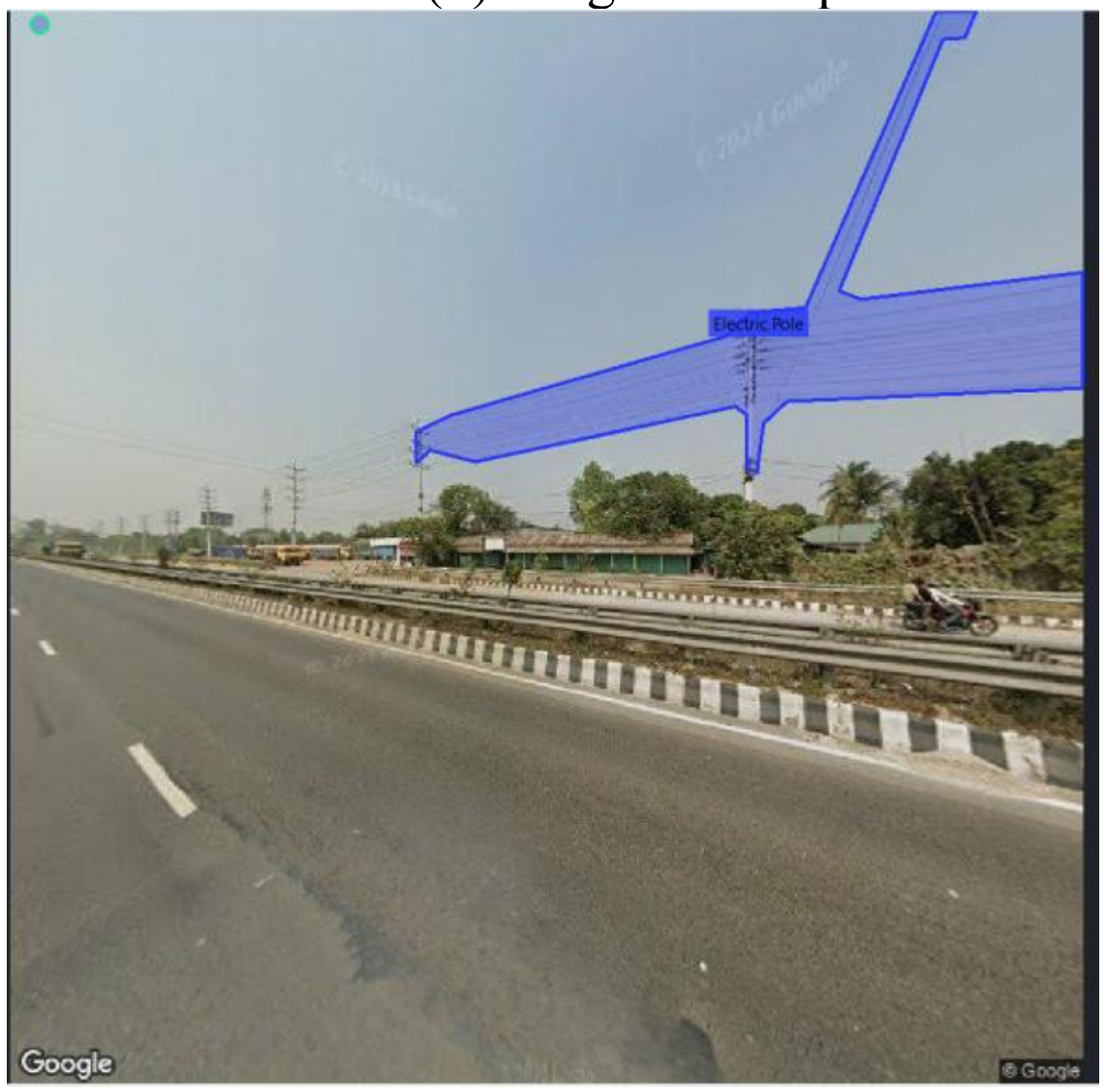

(d) Vertically Long Shape

**Figure 5** Different Features of Smart Selection based Selection and Annotation (mouse action-clicking)

**Figure 6** demonstrates model-suggested annotation using a pretrained SegFormer model trained on the Cityscapes dataset (Xie et al., 2021; Cordts et al., 2016). Unlike the annotation modes illustrated in Figures 3–5, this approach does not require the annotator to interact with individual objects during the initial segmentation stage. Instead, the model performs dense semantic prediction across the entire image in a single forward pass. The resulting class-probability maps are processed using a default softmax confidence threshold of 0.40. Connected components are then extracted separately for each predicted class, and components smaller than the default minimum area of 2,000 pixels are discarded. Each retained component is converted into a polygon suggestion and added to the annotation suggestion queue.

The annotator subsequently reviews the generated suggestions, accepting or rejecting each polygon and correcting its class assignment where necessary. This batch-oriented procedure differs from object-by-object annotation because the model initially identifies multiple candidate regions across the image, allowing the annotator to review a collection of proposals generated from a single inference operation. In the current implementation, the suggestions are presented with a provisional class assignment of index 0, rather than directly adopting the predicted semantic class, to account for potential classification errors arising from domain differences between Cityscapes and the target road-scene imagery. The annotator is therefore required to verify and assign the appropriate project-specific class before accepting each suggestion. This distinction is important because the predicted semantic category and the final annotation category are not necessarily equivalent.

The approach is particularly suitable for images containing multiple instances of road-scene categories represented in the pretrained model's taxonomy, such as roads, sidewalks, buildings, vegetation, and vehicle classes. In such cases, a single inference pass can generate multiple candidate annotations, potentially reducing the amount of manual interaction required per image. The extent of this efficiency gain, however, depends on the correspondence between the pretrained model's class taxonomy and the target annotation scheme, as well as the accuracy of its predictions under the target domain conditions (Liu et al., 2023).

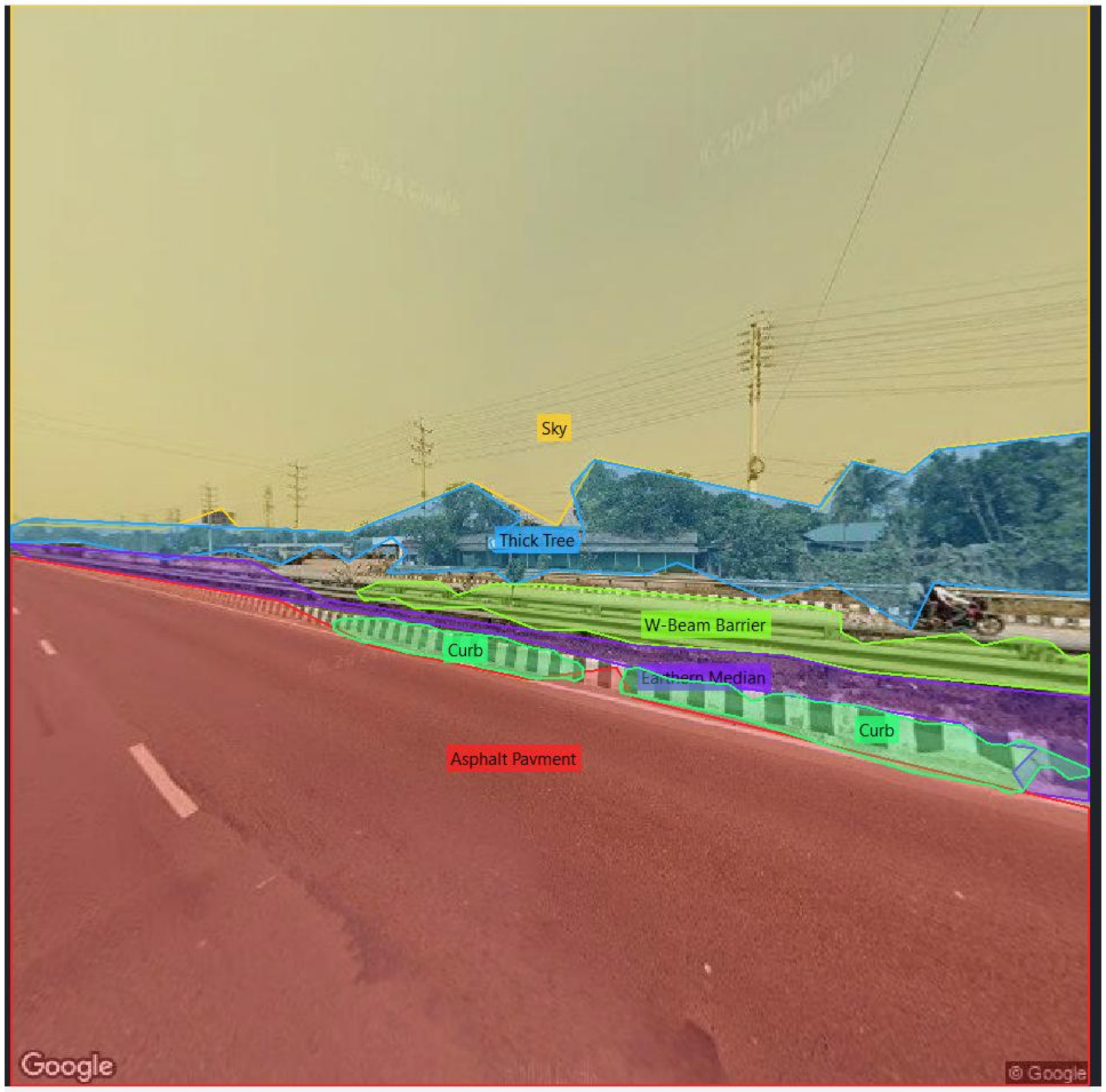

**Figure 6** Different Features of Model Suggested Selection and Annotation (no mouse action)

# 6 Results and Discussion

This section presents a qualitative assessment of the four annotation modes integrated into RoadAnnoter, using representative Google Street View images collected from recorded road accident locations in Bangladesh. The visual outputs are examined in terms of their ability to represent different road-scene features, capture object boundaries, and support annotation across diverse geometric configurations. A further visual comparison is presented between RoadAnnoter and the smart annotation functionality of Roboflow. No quantitative accuracy metrics are reported because a held-out reference dataset containing manually verified ground-truth polygons is not available at this stage. Accordingly, the findings and comparisons presented in this section are based exclusively on visual inspection of the generated annotations and should be interpreted as qualitative observations rather than quantitative evidence of segmentation accuracy or comparative superiority.

**Figure 7** presents representative outputs from the four annotation modes applied to different road-scene images from the study dataset. Each panel illustrates the annotation characteristics of one mode, highlighting its interaction requirements, feature coverage, and ability to represent different road and roadside geometries.

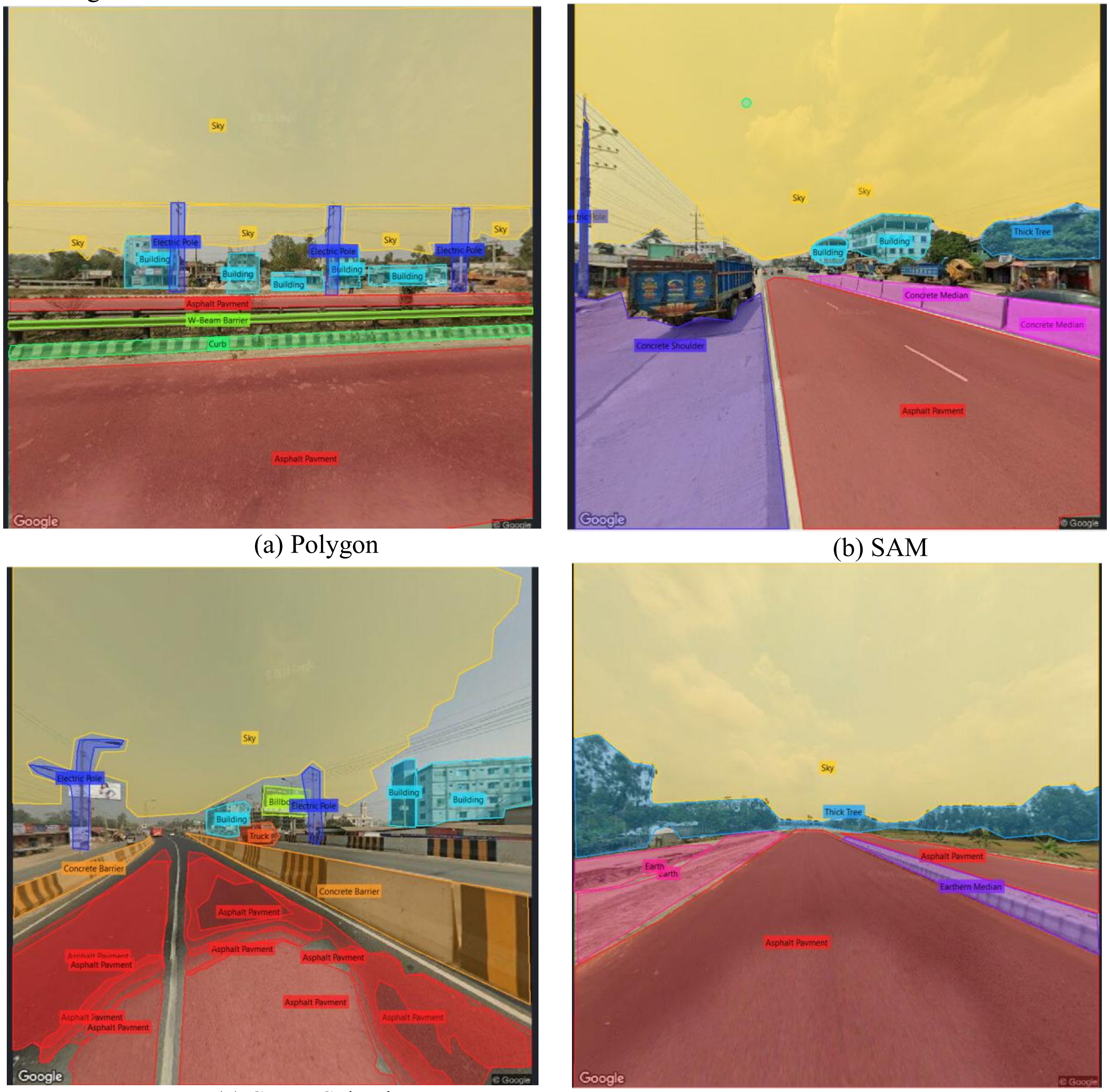

(a) Polygon (b) SAM

(c) Smart Selection (c) Model Predicted Selection

**Figure 7** Comparison of Different Selection and Annotation Methods

**Figure 7(a)** illustrates manual polygon annotation applied to a divided-highway scene containing complex overhead infrastructure. The annotator has delineated several road inventory classes, including sky (yellow), H-frame electric poles (cyan), roadside buildings (blue), asphalt pavement (red), a W-beam safety barrier (green), and a curb. Each feature is represented by a manually constructed polygon whose vertices are positioned according to the observed object boundaries. This example demonstrates the flexibility of manual annotation in accommodating arbitrary object geometries and project-specific classes, including complex structural elements such as cross-shaped pole frames that may not be represented as distinct categories in the pretrained models integrated into the application. However, the level of geometric detail achievable through this approach depends on the number and precision of manually placed vertices. Consequently, annotation of intricate structures can require substantially more interaction than the model-assisted alternatives. Manual polygon drawing is therefore particularly relevant when the target feature is geometrically complex, domain-specific, or inadequately represented by the available pretrained models.

**Figure 7(b)** presents the output of SAM Prompt on a road scene containing a concrete shoulder, an asphalt carriageway, a concrete median strip, dense roadside trees, and sky. The generated polygons follow the principal boundaries of the annotated regions, including the transitions between the carriageway and shoulder and between the road surface and surrounding vegetation. The shared mask-to-polygon conversion procedure, which includes morphological smoothing, Gaussian filtering, and Douglas–Peucker simplification, produces relatively smooth and compact polygon representations. This characteristic is useful for road inventory annotation, particularly when the target features comprise large, contiguous regions with visually identifiable boundaries. The example also illustrates the flexibility of iterative point prompting, through which the annotator can refine the proposed segmentation when the initial mask does not adequately represent the intended region.

**Figure 7(c)** demonstrates Smart Selection annotation in a grade-separated interchange scene containing multiple carriageway levels, concrete safety barriers, overhead structures, and sky. The single-click interaction generates polygon suggestions for broad asphalt pavement regions across different elevation levels, together with roadside infrastructure and other visible features. The example illustrates the potential of the one-click interaction strategy to reduce the number of prompting operations required for geometrically complex road scenes. Nevertheless, each distinct target region still requires an individual click, and segmentation near overpass and underpass boundaries requires careful review because the point-and-box prompt may include visually adjacent surfaces. Thus, although Smart Selection reduces interaction requirements relative to iterative prompting, the reliability of the resulting polygon remains dependent on the spatial context of the selected point and the candidate mask generated by SAM.

**Figure 7(d)** presents the output of the pretrained SegFormer B5 model trained on Cityscapes, applied to a straight rural single-carriageway scene. The model generates semantic predictions across the entire image in a single inference pass, from which polygon suggestions are extracted for the dominant scene regions. The illustrated outputs include sky and dense tree cover (cyan), asphalt pavement (red), a bare-earth shoulder (pink), and an earthen median (purple). The generated polygons provide broad spatial representations of these regions without requiring individual mouse interactions for each object. This example illustrates the potential efficiency of full-image semantic segmentation when the dominant scene categories are sufficiently represented in the pretrained model's taxonomy. However, the correspondence between predicted regions and the final project-specific annotation classes must still be verified by the annotator, particularly where differences between the Cityscapes training domain and the target imagery may affect semantic predictions. The visually smooth boundaries in this example reflect the combined effects of the model-generated masks and the subsequent mask-to-polygon processing stages; they should not, in the absence of reference annotations, be interpreted as evidence of calibrated confidence or quantitatively established boundary accuracy.

To provide an additional point of comparison, **Figure 8** presents polygon annotations generated using the smart annotation functionality of Roboflow on a comparable rural road scene from the same Google Street View dataset. The example is intended to illustrate the visual characteristics of polygon outputs produced by an external annotation platform under a similar road-scene context.

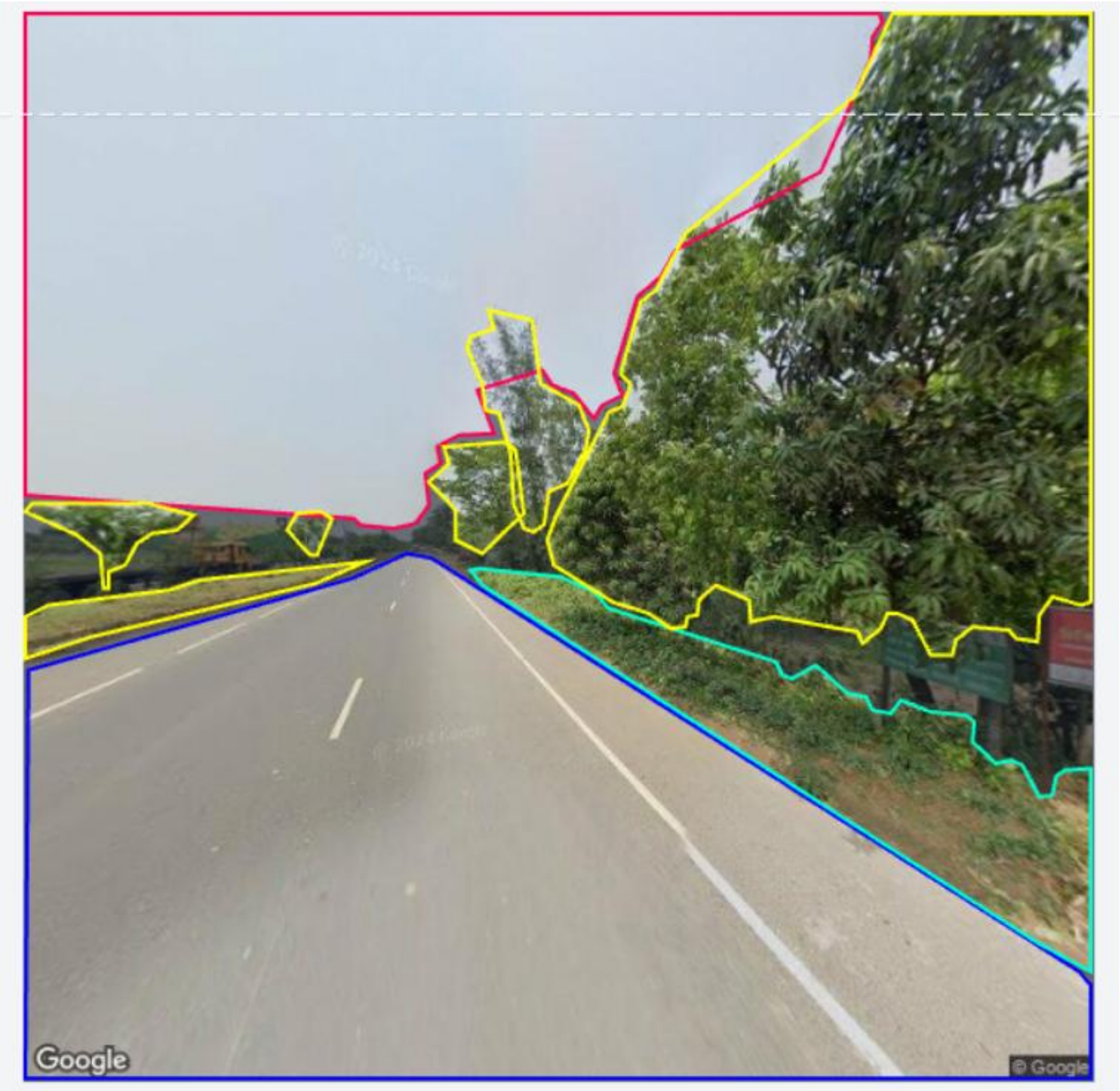

**Figure 8** Smart Select Annotation from Roboflow

**Figure 8** illustrates Roboflow's smart annotation output for a two-lane rural road scene comprising an asphalt carriageway, dense roadside tree cover on the right, and open sky. The resulting polygon boundaries are displayed using different outline colours, including blue for the road carriageway, red for the sky, and yellow and cyan for roadside vegetation. The road and sky polygons follow the principal region boundaries, while the vegetation polygons exhibit a comparatively irregular contour profile that captures variations in the visible foliage. Such contour characteristics may reflect differences in the segmentation and polygon-generation procedures, although the specific processing stages used to generate the displayed Roboflow output are not established from this visual comparison alone.

The RoadAnnoter outputs generated through SAM Prompt and Smart Selection exhibit visually smooth polygon boundaries on the illustrated road scenes. This appearance is partly attributable to the shared mask-to-polygon processing pipeline, which applies morphological closing, Gaussian blurring, and Douglas–Peucker simplification (Section 3.6). Roboflow's output also represents the principal scene regions with visually identifiable polygon boundaries. However, the examples do not establish that either tool achieves higher boundary accuracy, since the comparison is not based on identical images, controlled annotation conditions, or reference polygons. A rigorous comparative evaluation would require both tools to annotate the same images using identical class definitions, followed by assessment against a common ground-truth dataset using suitable polygon agreement measures, such as intersection over union (IoU) and boundary-based metrics.

The principal distinction demonstrated by the comparison is therefore functional rather than quantitative. RoadAnnoter integrates several complementary annotation approaches within a single desktop application, including manual polygon drawing for fine-grained and domain-specific features, Edge Fill for classical image-processing-based annotation in CPU-only environments, SAM-based interactive segmentation, and pretrained semantic segmentation for generating batch suggestions. Its locally operated workflow also supports offline annotation after the required model weights have been downloaded and made available locally. By contrast, Roboflow provides a separate annotation platform with its own smart annotation functionality. These differences highlight RoadAnnoter's emphasis on combining multiple annotation strategies and supporting locally managed road-scene annotation workflows, rather than establishing superior segmentation performance over an existing platform.

Taken together, the four annotation modes address complementary requirements in road safety image annotation. Manual polygon drawing provides the greatest flexibility for features with complex geometries or project-specific definitions that are not adequately represented by pretrained model taxonomies. SAM Prompt offers interactive control through iterative foreground and background prompting, making it suitable for visually distinguishable objects and surface regions that require segmentation refinement. Smart Selection prioritises interaction speed by generating a candidate segmentation from a single point-and-box prompt, particularly for relatively large and spatially coherent features. Pretrained SegFormer suggestions, in turn, support image-level batch proposal generation when the dominant scene categories correspond sufficiently well to the model's training taxonomy. The human-in-the-loop design provides a common quality-control mechanism across these approaches: generated suggestions are reviewed by the annotator and are not committed to the annotation files without explicit acceptance. Consequently, the system combines automated or semi-automated polygon generation with human verification, allowing the annotator to retain control over the final annotation output.

## 7 Conclusions

This study presents RoadAnnoter v1.0, a self-contained, open-source desktop application developed for polygon-based semantic segmentation annotation of road-scene images. The application integrates six complementary annotation modes within a unified human-in-the-loop framework: manual polygon drawing, SAM Prompt, Smart Selection, Edge Fill, pretrained semantic segmentation suggestions using Cityscapes-trained SegFormer models, and model-generated suggestions using the built-in TinySegNet architecture. The different modes accommodate varying levels of automation, computational requirements, and annotation complexity. Their outputs are processed through a common mask-to-polygon conversion pipeline incorporating morphological smoothing, Gaussian blurring, Suzuki–Abe contour extraction, optional edge-guided vertex refinement, and Douglas–Peucker simplification. The resulting annotations are stored in the YOLO segmentation format, facilitating their subsequent use in standard computer vision model-training workflows.

RoadAnnoter addresses several practical limitations associated with existing annotation workflows by combining classical image processing, promptable segmentation, pretrained semantic segmentation, and a built-in model-training capability within a single desktop environment. The integration of Cityscapes-trained SegFormer models provides pretrained assistance for commonly represented road-scene categories, including roads, sidewalks, vehicles, vegetation, buildings, and sky. However, the availability of these categories does not eliminate the need for annotator verification, particularly when differences between the pretrained model's taxonomy and the target annotation scheme introduce classification uncertainty. The Edge Fill mode provides an alternative annotation-assistance mechanism based entirely on classical image processing, allowing its use on CPU-only workstations without requiring deep learning models. Furthermore, the TinySegNet integration establishes a workflow through which accumulated project annotations can be used to train or retrain a lightweight segmentation model and subsequently generate additional annotation suggestions. This capability allows the annotation process to evolve alongside the development of a project-specific labelled dataset.

The qualitative assessment using Google Street View imagery collected from recorded road accident locations in Bangladesh demonstrated the complementary applicability of the annotation modes to different road-scene characteristics. Manual polygon drawing accommodated fine-structured and domain-specific features, including H-frame electric poles and W-beam barriers, for which direct manual delineation may be necessary. SAM Prompt provided interactive segmentation through foreground and background point prompts, allowing the annotator to refine candidate masks for visually distinguishable surface regions. Smart Selection reduced the interaction requirements by generating candidate masks from a single point-and-box prompt, facilitating the annotation of relatively large and spatially coherent features. Pretrained SegFormer models generated multiple semantic region proposals across an image in a single inference pass, providing a batch-oriented alternative when the dominant scene categories corresponded sufficiently well to the pretrained taxonomy. The visual comparison with Roboflow further illustrated the polygon output characteristics of an external smart annotation platform. Although both tools produced visually identifiable boundaries for the principal road-scene regions in the examples examined, the comparison was not conducted under controlled conditions and therefore does not establish comparative segmentation accuracy or superiority.

Several limitations of the current implementation and its evaluation warrant consideration. First, the shared mask-to-polygon pipeline retains only the external contour of each input mask. Consequently, enclosed background regions, such as pothole interiors within an annotated road surface or openings between structural elements, cannot be represented explicitly and may be incorporated into the resulting polygon. Second, the TinySegNet training procedure currently evaluates mean intersection over union (mIoU) using training batches without a separate validation dataset. The resulting metric therefore indicates performance on the training data rather than providing an independent assessment of model generalization. Third, the present evaluation is qualitative and does not include measured annotation time, polygon agreement against manually verified reference annotations, or inter-annotator variability. Accordingly, the reported observations demonstrate the application's functional capabilities but do not establish quantitative improvements in annotation efficiency or segmentation accuracy.

Future development and evaluation should address these limitations through several extensions. First, a systematic quantitative evaluation should be conducted using a manually verified reference dataset, with annotation time measured separately for each annotation mode. Polygon agreement can be assessed using intersection over union (IoU) and boundary F-score, while the number of required interactions and polygon vertices can provide additional measures of annotation effort and geometric complexity. Such an evaluation would establish a more objective basis for comparing the annotation modes and identifying their suitability for different road-scene features. Second, the TinySegNet training framework should incorporate a held-out validation dataset to support independent performance assessment and more reliable checkpoint selection. Data augmentation strategies, including horizontal flipping, colour jitter, and random cropping, could also be investigated to improve model robustness, particularly during the early stages of dataset development when labelled samples are limited. Third, the polygon representation and conversion pipeline should be extended to preserve enclosed holes and, where necessary, multiple disconnected regions within a single annotation. Finally, fine-tuning pretrained segmentation architectures using road-scene imagery from South and Southeast Asian contexts could improve their applicability to environments characterised by different road geometries, pavement conditions, traffic compositions, lane-use patterns, and roadside infrastructure compared with those represented in Cityscapes.

RoadAnnoter provides an integrated annotation framework that combines manual control, classical image processing, prompt-based segmentation, pretrained semantic prediction, and project-specific model development within a locally operated desktop application. By bringing these complementary approaches into a common annotation and data-management workflow, the application offers a flexible foundation for developing labelled road-scene datasets for transportation engineering, road infrastructure assessment, and road safety research. Its broader contribution lies in providing an extensible, human-supervised annotation environment in which different levels of automation can be selected based on the characteristics of the target features and the available computational resources.

## Declaration of Conflicting Interests

The authors declared no potential conflicts of interest with respect to the research, authorship, and/or publication of this article.

## Code Availability

The source code of RoadAnnoter v1.0 is archived on Zenodo and freely available under an open-source licence at https://doi.org/10.5281/zenodo.24314190 (Haque, 2026). GitHub repository: https://github.com/NazmulHaqueNazim/RoadAnnoter

## Appendix

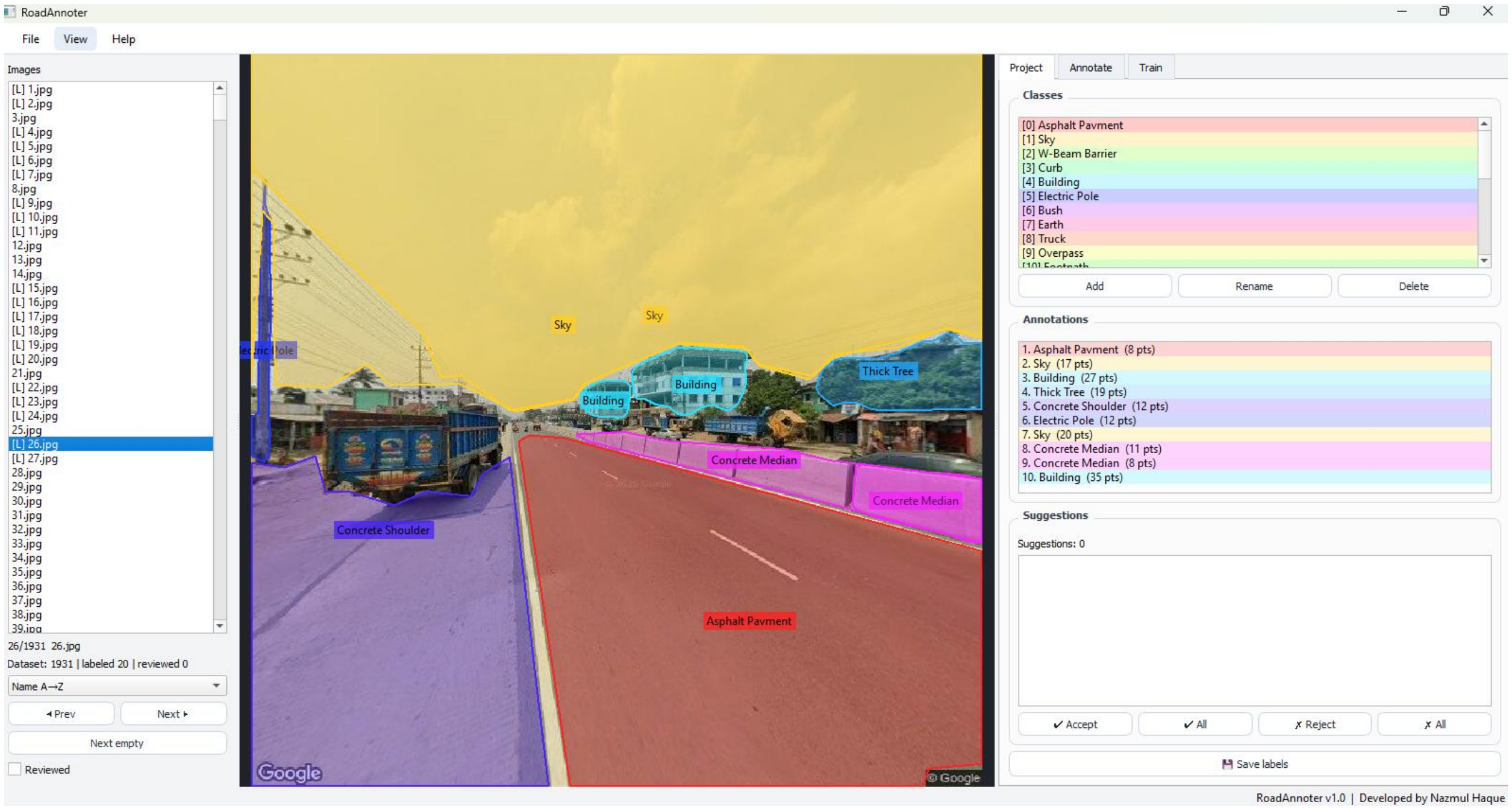

**Figure A1.** Snapshot of Annotation Procedure with Different Road Furniture Classes